\documentclass[conference]{IEEEtran}
\IEEEoverridecommandlockouts

\usepackage{cite}
\usepackage{amsmath,amssymb,amsfonts}
\usepackage{algorithmic}
\usepackage{graphicx}
\usepackage{textcomp}
\usepackage[center]{subfigure}
\usepackage{multirow}
\usepackage{bm}
\usepackage{siunitx}
\usepackage[
             pdfstartview=FitH,
             plainpages=false,
             bookmarksopen=true,
             bookmarksnumbered=true,
             colorlinks=true,
             bookmarkstype=toc]{hyperref} 

\usepackage{xcolor} 

\graphicspath{{graphics/},{logs/}}

\def\BibTeX{{\rm B\kern-.05em{\sc i\kern-.025em b}\kern-.08em
    T\kern-.1667em\lower.7ex\hbox{E}\kern-.125emX}}

\title{Active inference body perception and action \\for humanoid robots}

\author{%
Guillermo Oliver$^{1}$, Pablo Lanillos$^{1,2}$, Gordon Cheng$^{1}$
\thanks{$^{1}$Institute for Cognitive Systems, Technical University of Munich, Arcisstrasse 21, 80333 Munich, Germany.}%
\thanks{$^{2}$Donders Institute for Brain, Cognition and Behaviour, the Netherlands.}

\thanks{This work has been supported by SELFCEPTION project EU Horizon 2020 Programme, grant agreement n. 741941 and the EU's Erasmus+ Programme. Supplementary video: \url{https://youtu.be/rdbbmwo4TY4}, code: \url{tobereleased}.}
}

\begin{document}

\maketitle

\begin{abstract}

Providing artificial agents with the same computational models of biological systems is a way to understand how intelligent behaviours may emerge. We present an \textit{active inference} body perception and action model working for the first time in a humanoid robot. The model relies on the free energy principle proposed for the brain, where both perception and action goal is to minimise the prediction error through gradient descent on the variational free energy bound. The body state (latent variable) is inferred by minimising the difference between the observed (visual and proprioceptive) sensor values and the predicted ones. Simultaneously, the action makes sensory data sampling to better correspond to the prediction made by the inner model. We formalised and implemented the algorithm on the iCub robot and tested in 2D and 3D visual spaces for online adaptation to visual changes, sensory noise and discrepancies between the model and the real robot. We also compared our approach with classical inverse kinematics in a reaching task, analysing the suitability of such a neuroscience-inspired approach for real-world interaction. The algorithm gave the robot adaptive body perception and upper body reaching with head object tracking (toddler-like), and was able to incorporate visual features online (in a closed-loop manner) without increasing the computational complexity. Moreover, our model predicted involuntary actions in the presence of sensorimotor conflicts showing the path for a potential proof of active inference in humans.

\end{abstract}


\section{Introduction}
\label{sec:intro}

The medical doctor and physicist Hermann von Helmholtz described visual perception as an unconscious mechanism that infers the world \cite{Helmholtz1867}. In other words, the brain has generative models that reconstruct the world from partial information. Nowadays, there is a scientific mainstream that describes the inner workings of the brain as those of a Bayesian inference machine~\cite{knill2004bayesian,friston2005theory}. This approach supports that we can adjust the cues (visual, proprioceptive, tactile, etc.) contribution to our interpretation in a Bayesian optimal way considering sensors and motor uncertainties \cite{parr2017uncertainty}. This implies that the brain is able to encode uncertainty not only for perception but also for acting in the world.

\setlength{\belowcaptionskip}{-10pt}

\begin{figure}[t]
	\centering
	\includegraphics[width=0.88\linewidth, height=130px]{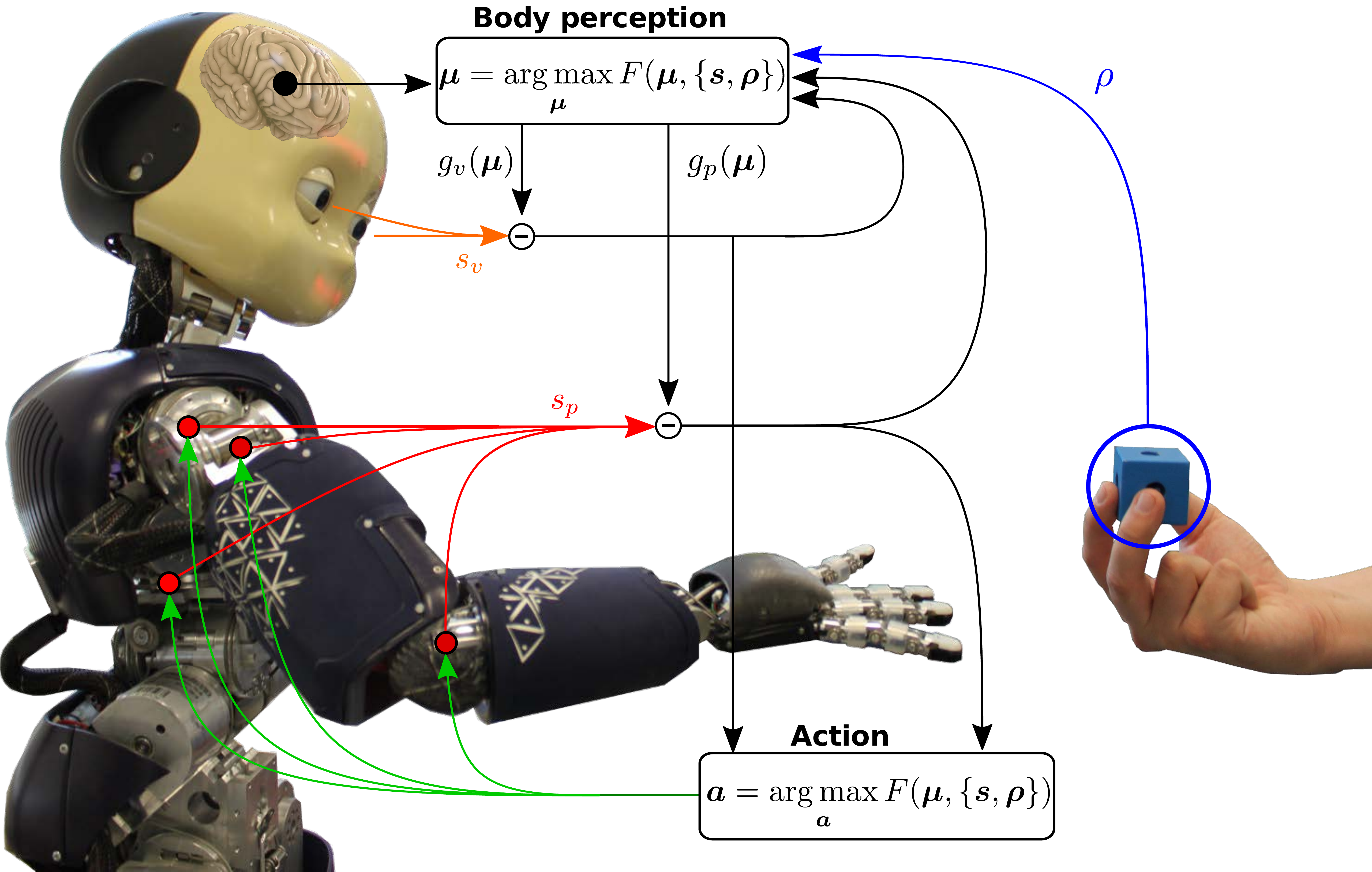}
	\caption{Body perception and action via \textit{active inference}. The robot infers its body (e.g., joint angles) by minimising the prediction error: discrepancy between the sensors (visual $\bm{s_v}$ and joint $\bm{s_p}$) and their expected values ($\bm{g_v}(\bm{\mu})$ and $\bm{g_p}(\bm{\mu})$). In the presence of error it changes the perception of its body $\bm{\mu}$ and generates an action $\bm{a}$ to reduce this discrepancy. Both are computed by optimising the free energy bound $F$. In a reaching task, the object is a causal variable $\bm{\rho}$ that acts as a perceptual attractor, producing an error in the desired sensory state and promoting an action towards the goal. The equilibrium point $\bm{\dot{\mu}}=0$ appears when the hand reaches the object. The head keeps the object in its visual field, improving the reaching performance.
	}
	\label{img:setup}
	\vspace{-5mm}
\end{figure}
\setlength{\belowcaptionskip}{0pt}

Optimal feedback control was proposed for modelling motor coordination under uncertainty \cite{todorov2002optimal}. Alternatively, \textit{active inference} \cite{friston2010unified}, defended that both perception and action are two sides of the same process: the free energy principle. This principle accounts for perception, action and learning through the minimization of \textit{surprise}, i.e., the discrepancy between the current state and the predicted or desired one. According to this approach, free energy is a way of quantifying surprise and it can be optimised by changing the current beliefs (perception) or by acting on the environment (action) to adjust the difference between reality and prediction \cite{friston2010action}. Active inference would allow living beings to infer and simultaneously adapt their body to uncertain environments \cite{kirchhoff2018markov}, something just partially solved in robotics.

Hence, inspired by this principle, we present a robust and adaptive body perception and action mathematical model based on \textit{active inference}, for a real humanoid robot  (Fig.~\ref{img:setup}). Our approach describes artificial body perception and action, similar to biological systems, as a flexible and dynamic process that approximates the robot body latent state using the error between the expected and the observed sensory information. The proposed model enabled the robot to have adaptive body perception and to perform upper-body reaching behaviours even under high levels of sensor noise and discrepancies between the model and the real robot. We further compared our approach to standard inverse kinematics and analysed the relevance and the challenges of predictive coding interpretation solved through variational free energy methods for humanoid robotics. For reproducibility, the code, with the parameters used, is publicly available: \url{tobereleased}. 

\subsection{Related work}
\label{sec:related}

Multisensory body perception has been widely studied in the literature, from the engineering point of view, and enables robots to estimate its body state by combining joint information with other sensors such as images and tactile cues. Bayesian estimation (e.g., particle filters) has been proved to achieve robust and accurate model-based robot arm tracking \cite{fantacci2017visual} even under occlusion \cite{bohg2016}, or for simultaneous joint offset parameters adaptation and hand pose estimation \cite{vicente2016online}. Furthermore, integrated visuomotor processes enabled humanoid robots to learn motor representations for robust reaching \cite{gaskett2003online,jamone2014autonomous}, visuotactile motor representations for reaching and avoidance behaviours \cite{roncone2016peripersonal} or enabling visual robot body distinction for aiding the discovering of movable objects \cite{lanillos2016yielding}. Free energy optimization for multisensory perception in a real robot was firstly presented in \cite{lanillos2018adaptive} working as an approximate Bayesian filter estimation. The robot was able to perceive its arm location fusing visual, proprioceptive and tactile information.


Active inference (under the free energy principle) managed to include the action within the multisensory perception process, explained as a classical spinal reflex arc pathway triggered by prediction errors. It has been mainly studied in theoretical or simulated conditions. Friston et al. \cite{friston2010action} presented this neuroscience-inspired model as a generalization of the dynamic expectation-maximization algorithm. Recently, a few works analysed this approach for robotics in simulated conditions: one degree of freedom vehicle \cite{baltieri2017}, a PR2 robot arm \cite{pio2016active}, and two degrees of freedom robot arm with generative function learning \cite{lanillos2018active}. However, these studies did not validate the method on real robots bypassing the true difficulties, such as non-linear dynamics due to friction, backlash, unmodelled physical constraints, uncertainties in the dynamic model, sensor and model errors, etc.

In this work, we provide the active inference perception and action mathematical construct for the humanoid robot iCub upper-body working in different real setups. Thus, we show the plausibility of real implementations of the free energy principle in artificial agents, analysing its challenges and limitations. Furthermore, we introduce a novel way to compute the mapping from action to sensory consequences (partial derivatives of the sensory input with respect to the action) exploiting the properties of velocity control, and simplifying the calculus of the action within the free energy optimization framework.


\subsection{Paper organization}
First, in Sec. \ref{sec:math} we explain the general mathematical free energy optimization for perception and action. Afterwards, Sec. \ref{sec:icub} describes the iCub physical model and in Sec. \ref{sec:model} and \ref{sec:model:eye} we detail the active inference computational model applied for object reaching and head object tracking. Finally, Sec. \ref{sec:results} shows the results for several experiments and Sec. \ref{sec:discussion} analyses the relevance of implementing active inference body perception in humanoids, for both human and robotics science.

\section{Free energy optimization}
\label{sec:math}

\subsection{Bayesian inference}
\label{sec:math:bayes}

According to the Bayesian inference model for the brain, the body configuration $x$ is inferred using available sensory data $s$ by applying Bayes' theorem: $p(x|s) = p(s|x) p(x)/p(s)$. Where this \textit{posterior probability}, corresponding to the probability of body configuration $x$ given the observed data $s$, is obtained as a consequence of three antecedents: (1) \textit{likelihood}, $p(s|x)$, or compatibility of the observed data $s$ with the body configuration $x$, (2) \textit{prior probability}, $p(x)$, or current belief about the configuration before receiving the sensory data $s$, and (3) \textit{marginal likelihood}, $p(s)$, a normalization term which corresponds to the marginalization of the likelihood of receiving sensory data $s$ regardless of the configuration. 



The goal is to find the value of $x$ which maximises $p(x|s)$, because it is the most likely value for the real-world body configuration $x$ taking into account the sensory data obtained $s$. This direct method becomes intractable for large state spaces \cite{beal2002bayesian}. In this case, marginalization over all the possible body and world states explodes computationally.

\subsection{Free energy principle with the Laplace approximation}
\label{sec:math:free energy}

The \textit{free energy principle} \cite{friston2010unified} provides a tractable solution to this obstacle, where, instead of calculating the marginal likelihood, the idea is to minimise the \textit{Kullback-Leibler divergence} \cite{kullback1951} between a reference distribution $q(x)$ and the real $p(x|s)$, in order for the first to be a good approximation of the second.

\vspace{-3mm}

\begin{gather}
D_{KL}(q(x) || p(x|s)) = \int q(x) \ln \frac{q(x)}{p(x|s)} dx \nonumber \\
=\int q(x) \ln \frac{q(x)}{p(s,x)} dx + \ln p(s)
= F + \ln p(s) \geq 0
\end{gather}


Minimising the first term, $F$, effectively minimises the difference between these two densities, with only the marginal likelihood remaining. Unlike the whole expression of the KL-divergence, the first term can be evaluated because it depends on the reference distribution and the knowledge about the environment we can assume the agent has, $p(s,x) = p(s|x) p(x)$. This term is defined as \textit{variational free energy} \cite{buckley2017free}.




According to the free energy optimization theory, there are two ways to minimise \textit{surprise}, which accounts for the discrepancy between the current state and the predicted or desired one (prediction error): changing the belief (\textit{perceptual inference}) or acting on the world (\textit{active inference}). Perceptual inference and active inference optimise the value of variational free energy, while active inference also optimises the value of the marginal likelihood by acting on the environment and changing the sensory data $s$.

Under the Laplace approximation instead of computing the whole distribution we can track the mode \cite{friston2007variational} and simplify the calculus of $F$. Thus, the variational density $q(x)$ is assumed to have Gaussian form $\mathcal{N}(\bm{x} | \bm{\mu} ,\Sigma)$. Defining the Laplace-encoded energy \cite{buckley2017free} as $L(s, x)=-\ln p(s,x)$, $F$ can be approximated to:
\vspace{-3mm}

\begin{align}
F \approx L(s,\bm{\mu})  - \left[ \frac{1}{2} \ln | \Sigma | + n \ln 2\pi \right] 
\end{align}
where $\bm{\mu}$ conceptually describes the internal belief or state, $s$ is the sensory input and $n$ is the size of $\bm{\mu}$ vector.

\subsection{Perceptual and active inference}
\label{sec:math:optimization}

\textit{Perceptual inference} is the process of updating the inner model belief to best account for sensory data, minimising the prediction error. For instance for body perception, the agent must infer the most-likely or optimal value for the body configuration or state $\mu$. This optimal value is the one that minimises free energy. To solve it, we can compute it by gradient descent over the free energy term. For static systems, this update is performed directly as \cite{lanillos2018adaptive}: $\dot{\mu} = -\frac{\partial F}{\partial \mu}$. In dynamic systems, time derivatives should be considered, the state variable is now a vector $\bm{\mu}$:


\begin{equation}
\dot{\bm{\mu}} = \bm{D} \bm{\mu} - \frac{\partial F}{\partial \bm{\mu}}
\label{eq:gradient-free energy-internal-dynamic}
\end{equation}

\noindent where the $\bm{D}$ is the block-matrix derivative operator. When free energy is minimised, the value of its derivative is $\frac{\partial F}{\partial \mu} = 0$, and the system is at equilibrium. 





\textit{Active inference} \cite{friston2010action}, is the extension of perceptual inference to the relationship between sensors and actions, taking into account that actions can change the world to make sensory data more accurate with predictions made by the inner model. The action plays a core role in the optimization and improves the approximation of the real distribution, therefore reducing the prediction error by minimising free energy. It also acts on the marginal likelihood by changing the real configuration which modifies the sensory data $s$ to obtain new data that is more in concordance with the agent's belief. The optimal value is the one which minimises free energy, and again a gradient descent approach will be taken to update its value:

\begin{equation}
\dot{a} = -\frac{\partial F}{\partial a}
\label{eq:gradient-free energy-action}
\end{equation}

 

\section{Robot physical model}
\label{sec:icub}

\begin{figure}[h!]
	\centering
	\includegraphics[width=0.9\linewidth]{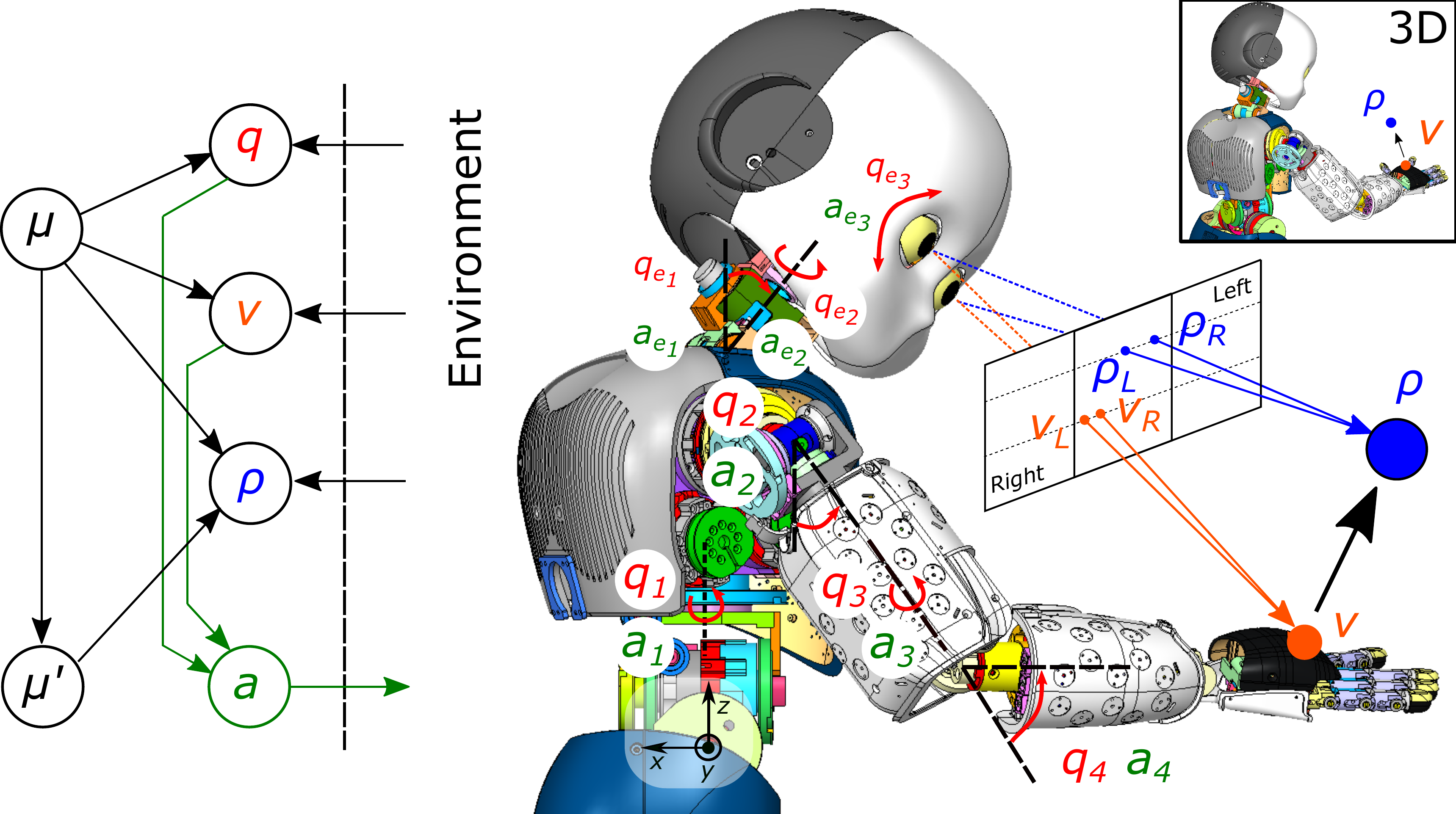}
	\caption{Model description. (Left) Generative model of the robot. (Right) Notation and robot configuration. Shown variables: internal variables $\bm{\mu}$ and $\bm{\mu'}$, joint angle position $\bm{q}$, joint actions $\bm{a}$, and 3D location of the end-effector $\bm{v}$ and causal variables $\bm{\rho}$ along with their projections in 2D visual space.}
	\label{fig:model}
\end{figure}

The robot iCub \cite{iCub2008} (v1.4) is a humanoid powered by electric motors and driven by tendons. It is divided into several kinematic chains defined through homogeneous transformation matrices using Denavit-Hartenberg convention. We focused on two kinematic chains, those with the end-effector being the hands (without fingers), the head, the eyes and the torso. 


As depicted in Fig. \ref{fig:model}, each arm model is defined as a three degree of freedom system: \textit{shoulder\_roll}, \textit{shoulder\_yaw} and \textit{r\_elbow}. To provide improved reaching capabilities, an additional degree of freedom is considered for the torso: \textit{torso\_yaw}. Left and right eye cameras observe the end-effector position and the world around it, providing stereo vision. The joints considered for the head are: \textit{neck\_pitch} and \textit{neck\_yaw}. Moreover, vertical synchronised motion is performed in both eyes using: \textit{eyes\_tilt}.



\section{Active inference computational model for iCub reaching task}
\label{sec:model}


\subsection{Problem formulation}
\label{sec:model:single:proprio}

The body configuration, or internal variables, is defined as the joint angles $\bm{g_p}(\bm{\mu}) = \bm{\mu}$. The estimated states $\bm{\mu}\in\rm{I\!R^4}$ are the belief the agent has about the joint angle position and the action $\bm{a}\in\rm{I\!R^4}$ is the angular velocity of those same joints. Due to the fact that we use a velocity control for the joints, first-order dynamics must also be considered $\bm{\mu'}\in\rm{I\!R^4}$.


Sensory data will be obtained through several input sensors. Using binocular disparity between the images obtained from both eyes, stereo vision reconstruction is performed to obtain the 3D position of the end-effector, $\bm{s_v}\in\rm{I\!R^3}$. All joint motors have encoders which provide joint angle position, $\bm{s_p}\in\rm{I\!R^4}$. 

\small
\begin{equation}
\bm{s_p} = \left[\begin{array}{c} 
q_1 \\
q_2 \\
q_3 \\
q_4 \\
\end{array} \right] \qquad
\bm{s_v} = \left[\begin{array}{c} 
v_1 \\
v_2 \\
v_3 \\
\end{array} \right]
\end{equation}
\normalsize

The likelihood $p(\bm{s}|\bm{\mu})$ is made up of proprioception functions in terms of the current body configuration, while the prior $p(\bm{\mu})$ will take into account the dynamic model of the agent that describes how this internal state changes with time. Adapting the Laplace-encoded energy for our robot model and assuming independence between different modalities sensory readings:

\vspace{-10pt}

\begin{equation}
\ln p(\bm{s},\bm{\rho}, \bm{\mu}) = \ln  p(\bm{s_p}|\bm{\mu}) p(\bm{s_v}|\bm{\mu}) p(\bm{\mu'}|\bm{\mu}, \bm{\rho})
\label{eq:neg-free energy-term}
\end{equation}

\subsection{Free energy optimization}
\label{sec:model:single:free energy}

In order to define the conditional densities for each of the terms, we should define the expressions for the sensory data. Joint angle position, $\bm{s_p}$, is obtained directly from the joint angle sensors. Let us assume that the input is noisy and follows a normal distribution with mean at the internal value $\bm{\mu}$ and variance $\Sigma_{s_p}$. End-effector 2D or 3D position, $\bm{s_v}$, is defined by a non-linear function dependent on the body configuration and obtained using the forward model of the arm. Again, the input is noisy and follows a normal distribution with mean at the value of this function $\bm{g_v}(\bm{\mu})\in\rm{I\!R^{n_v}}$ and variance $\Sigma_{s_v}$, where $n_v$ is the visual dimensions. The dynamic model for $n_p$ latent variables (joints angles) is determined by a function which depends on both the current state $\bm{\mu}$ and the causal variables $\bm{\rho}$ (e.g. 3D position of the object to be reached), with a noisy input following a normal distribution with mean at the value of this function $\bm{f}(\bm{\mu},\bm{\rho})\in\rm{I\!R^{n_p}}$ and variance $\Sigma_{s_\mu}$. Therefore, likelihood functions in Eq. \eqref{eq:neg-free energy-term} are defined as:


\small
\begin{gather}
p(\bm{s_p}|\bm{\mu}) = \prod_{i=1}^{n_p} \mathcal{N}(q_i|\mu_i,\Sigma_p) \qquad
p(\bm{s_v}|\bm{\mu}) = \prod_{i=1}^{n_v} \mathcal{N}(v_i|g_{v_i}(\bm{\mu}),\Sigma_v) \nonumber \\
p(\bm{\mu'}|\bm{\mu},\bm{\rho}) = \prod_{i=1}^{n_p} \mathcal{N}(\mu'_i|f_i(\bm{\mu},\bm{\rho}),\Sigma_{s_{\mu}})
\end{gather}
\normalsize




The optimization of the Laplace-encoded energy is performed using gradient descent (Eq. \eqref{eq:gradient-free energy-internal-dynamic} and \eqref{eq:gradient-free energy-action}). The dependency of $F$ with respect to the vector of internal variables $\bm{\mu}$ can be calculated using the chain rule on the functions that depend on those internal variables. The dependency of $F$ with respect to the vector of actions $\bm{a}$ is calculated considering that the only magnitudes directly affected by action are the values obtained from the sensors: $\frac{\partial F}{\partial a} = \frac{\partial s}{\partial a} \frac{\partial F}{\partial s}$ - see Sec. \ref{sec:model:ai}.

\small
\begin{align}
-\frac{\partial F}{\partial \bm{\mu}}
&=  \frac{1}{\Sigma_{s_p}}(\bm{s_p} - \bm{\mu})
+ \frac{1}{\Sigma_{s_v}} \frac{\partial \bm{g_v}(\bm{\mu})}{\partial \bm{\mu}}^T (\bm{s_v} - \bm{g}(\bm{\mu})) \nonumber \\
& + \frac{1}{\Sigma_{s_{\mu}}} \frac{\partial \bm{f}(\bm{\mu},\bm{\rho})}{\partial \bm{\mu}}^T (\bm{\mu'} - \bm{f}(\bm{\mu},\bm{\rho})) 
\label{eq:multi-perception}\\
-\frac{\partial F}{\partial \bm{a}}
&=  - \left(\frac{1}{\Sigma_{s_p}} \frac{\partial \bm{s_p}}{\partial \bm{a}}^T (\bm{s_p} - \bm{\mu}) 
+ \frac{1}{\Sigma_{s_v}} \frac{\partial \bm{s_v}}{\partial \bm{a}}^T (\bm{s_v} - \bm{g_v}(\bm{\mu})) \right)
\label{eq:multi-action}
\end{align}
\normalsize

The agent also infers the first-order dynamics of the body $\bm{\mu'}$ and updates their values using a gradient descent formulation. The dependency of $F$ with respect to $\bm{\mu'}$ is limited to the influence of the dynamic model.

\small
\begin{equation}
-\frac{\partial F}{\partial \bm{\mu'}}
= -\frac{1}{\Sigma_{s_{\mu}}} (\bm{\mu'} - \bm{f}(\bm{\mu},\bm{\rho})) = \frac{1}{\Sigma_{s_{\mu}}} (\bm{f}(\bm{\mu},\bm{\rho})- \bm{\mu'})
\label{eq:multi-first-order}\\
\end{equation}
\normalsize

Finally, the differential equations for $\bm{\mu}$, $\bm{\mu'}$ and $\bm{a}$ are:

\begin{gather}
\dot{\bm{\mu}} = \bm{\mu'} - \frac{\partial F}{\partial \bm{\mu}} \qquad
\dot{\bm{\mu'}} = -\frac{\partial F}{\partial \bm{\mu'}} \qquad
\dot{\bm{a}} = -\frac{\partial F}{\partial \bm{a}}
\label{eq:update-equations}
\end{gather}
A first-order Euler integration method is applied to update the values of $\bm{\mu}$, $\bm{\mu'}$ and $\bm{a}$ in each iteration. E.g., $\mu_{i+1} = \mu_{i} + \dot{\mu} \, \Delta t$, where $\Delta t = T$ is the period of execution of the updating cycle.

\subsection{Perceptual attractor dynamics}
\label{sec:model:attractor}

The reaching goal is defined in the dynamics of the model by introducing a perceptual attractor $\bm{\rho}$ in the 2D or 3D space:

\begin{equation}
\bm{A}(\bm{\mu},\bm{\rho}) = \rho_4
\left(\left[
\begin{array}{c}
\rho_1 \\
\rho_2 \\
\rho_3
\end{array} \right]  - \left[
\begin{array}{c}
g_{v_1}(\bm{\mu}) \\
g_{v_2}(\bm{\mu}) \\
g_{v_3}(\bm{\mu})
\end{array} \right] \right)
\label{eq:armattractor}
\end{equation}

where $\rho_{1:3}$ defines the object location in the visual space and $\rho_{4}$ describes the gain of the attractor.

Internal variable dynamics are then defined in terms of the attractor: $\bm{f}(\bm{\mu},\bm{\rho}) = \bm{T}(\bm{\mu}) \bm{A}(\bm{\mu},\bm{\rho})$. $\bm{T}(\bm{\mu})$ is the function that transforms the attractor vector from target space (3D) to joint space. The system is velocity controlled, therefore target space is a linear velocity vector and joint space is angular velocity.



When the Jacobian matrix is rectangular (e.g., 3D space and 4 DOF arm joint space), we used the generalised inverse (Moore-Penrose pseudoinverse) as the mapping function:  $\bm{T}(\bm{\mu}) = \bm{J}^{+}(\bm{\mu})$. This matrix is calculated using the singular-value decomposition (SVD), where $\bm{J}^{+} = \bm{V} \bm{\Sigma}^{+} \bm{U}^T$.

\subsection{Active inference}
\label{sec:model:ai}
Action is set to be an angular velocity magnitude, which corresponds with angular joint velocity in the latent space. In order to compute Eq. \eqref{eq:multi-action} we need the mapping between the sensor values and the velocities of the joints, expressed in the partial derivatives $\frac{\partial \bm{s_p}}{\partial \bm{a}}$ and $\frac{\partial \bm{s_v}}{\partial \bm{a}}$.  


We assume that the control action, $a$, is updated for every cycle, and therefore for each interval of time between cycles it has the same value. For each period (cycle time between updates), the equation of uniform circular motion is satisfied for each joint position. If this equation is discretised, for each sampling moment, which are $T$ seconds apart, the joint value will be updated in the following way: $q_{i+1} = q_i + a_i \; T$. The dependency of the joint angle position with respect to the control action is therefore defined.





The partial derivatives of joint position $\bm{s_p}$ with respect to action, considering there is no cross-influence or coupling between joint velocities and that $q_i$ and its expected $\mu_i$ should converge at equilibrium, are given by the following expression:

\begin{equation}
\frac{\partial q_i}{\partial a_j} = \frac{\partial \mu_i}{\partial a_j} = 
\begin{cases} T & i = j, \\
0 &\text{otherwise}. 
\end{cases}
\end{equation}


 
If the dependency of joint position with respect to action is known, we can use the chain rule to calculate the dependency for the visual sensor $\bm{s_v}$. Considering that the values of $g_{v_i}(\bm{\mu})$ should also converge to $v_i$ at equilibrium, the partial derivatives are given by the following expression:

\begin{equation}
\frac{\partial v_i}{\partial a_j} = \frac{\partial v_i}{\partial q_j} \frac{\partial q_j}{\partial a_j} = \frac{\partial g_{v_i}(\bm{\mu})}{\partial \mu_j} \frac{\partial \mu_j}{\partial a_j}
\end{equation}

\section{Active inference computational model for iCub head object tracking task}
\label{sec:model:eye}

We extend the arm reaching model for the head to obtain an object tracking motion behaviour. The goal of this task is to maintain the object in the centre of the visual field, thus increasing its reaching working range capabilities. 

Sensory data and proprioception for the head is defined by internal variables beliefs $\bm{\mu_e}\in\rm{I\!R^3}$, actions $\bm{a_e}\in\rm{I\!R^3}$, and first-order dynamics $\bm{\mu'_e}\in\rm{I\!R^3}$ and because the end-effector are the eyes, there is only joint angle position, $\bm{s_e}\in\rm{I\!R^3}$.

\begin{figure*}[t!]
	\centering
	\subfigure[Sensory fusion path.]{
		\centering
		\includegraphics[width=0.3\textwidth]{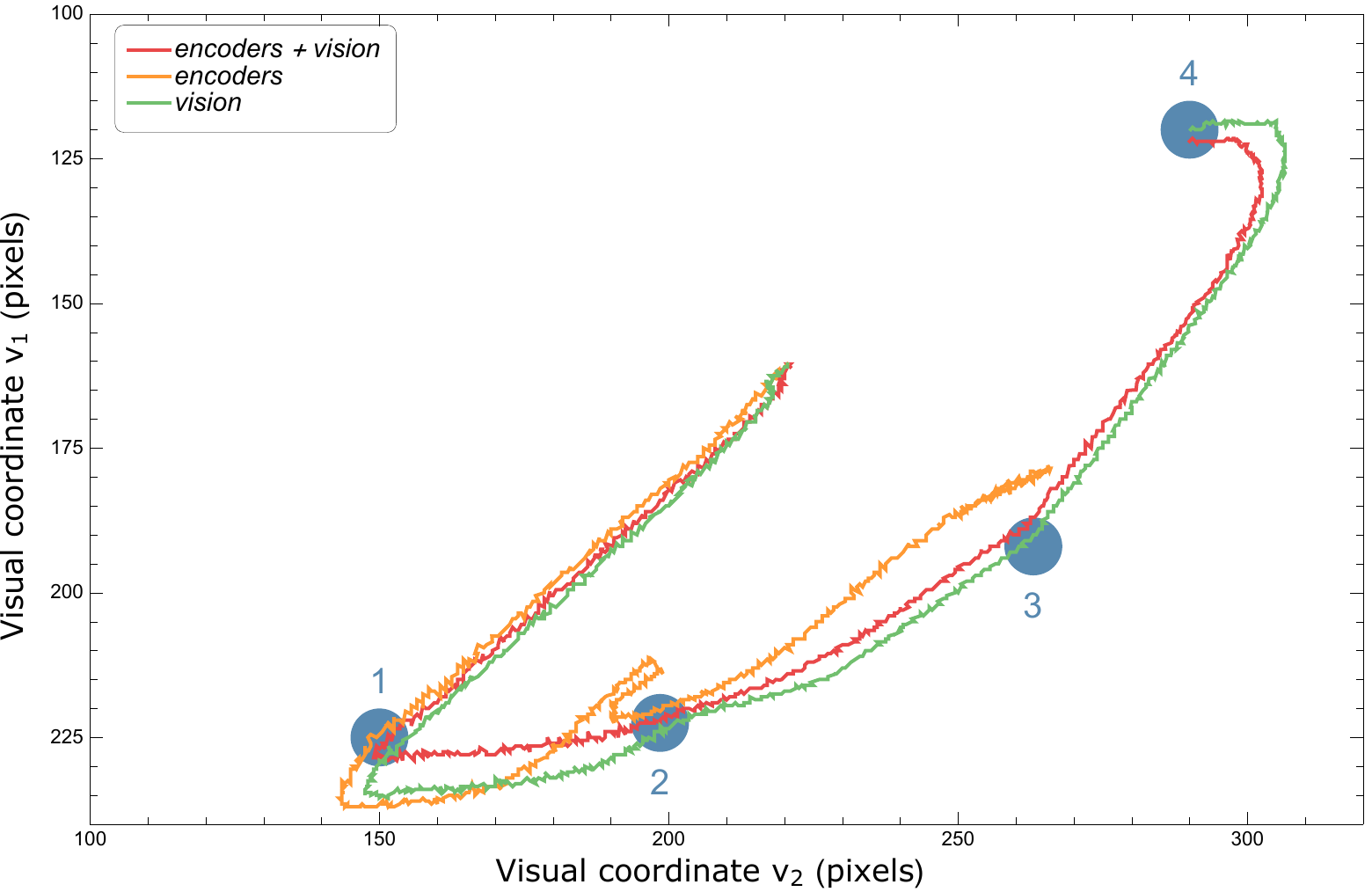}
		\label{results:poc:path-cycle}
	}
	\subfigure[Visual marker deviation to the left.]{
		\centering
		\includegraphics[width=0.3\textwidth]{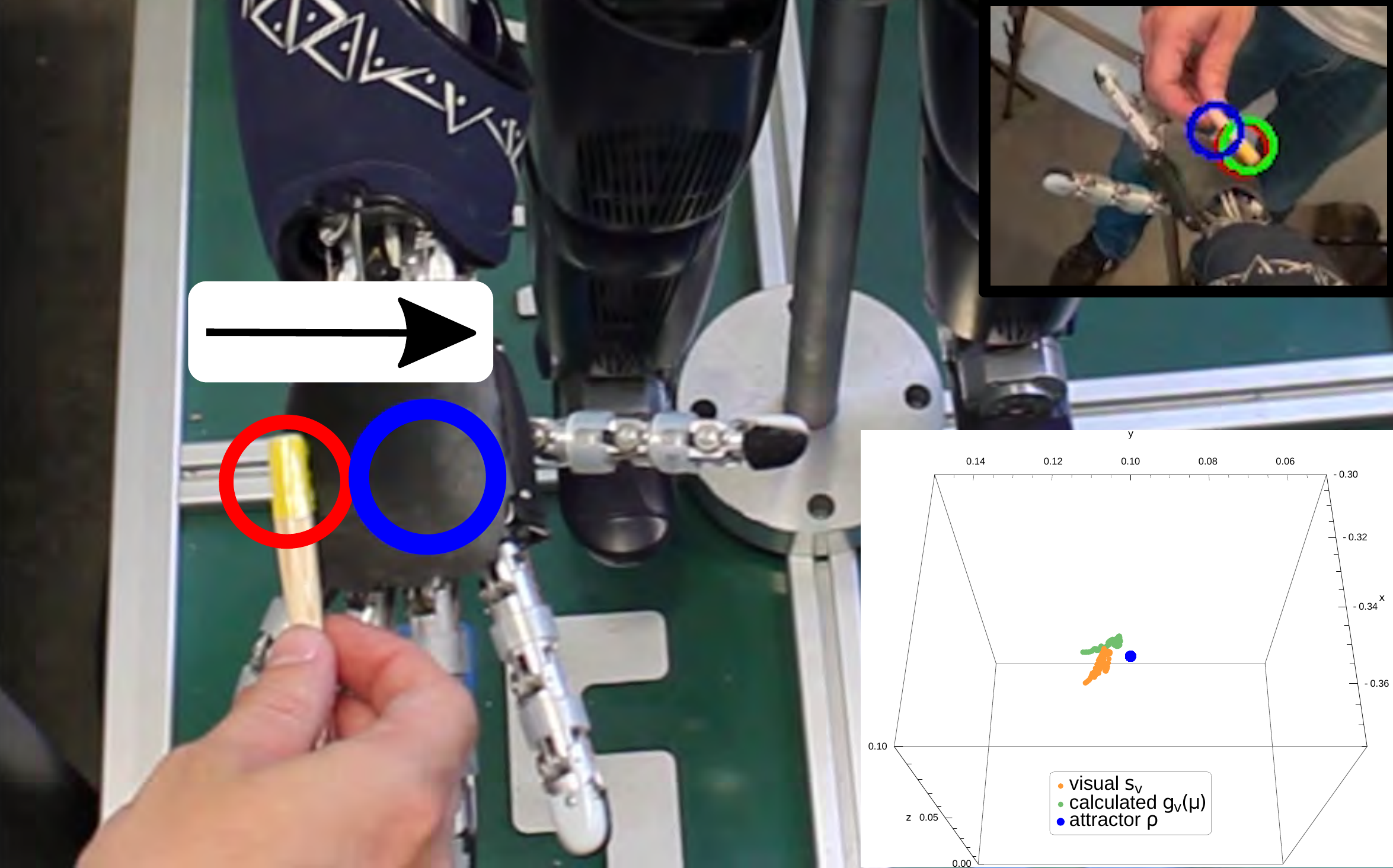}
		\label{results:adapt:left-deviation}
	}	
	\subfigure[Visual marker deviation to the right.]{
		\centering
		\includegraphics[width=0.3\textwidth]{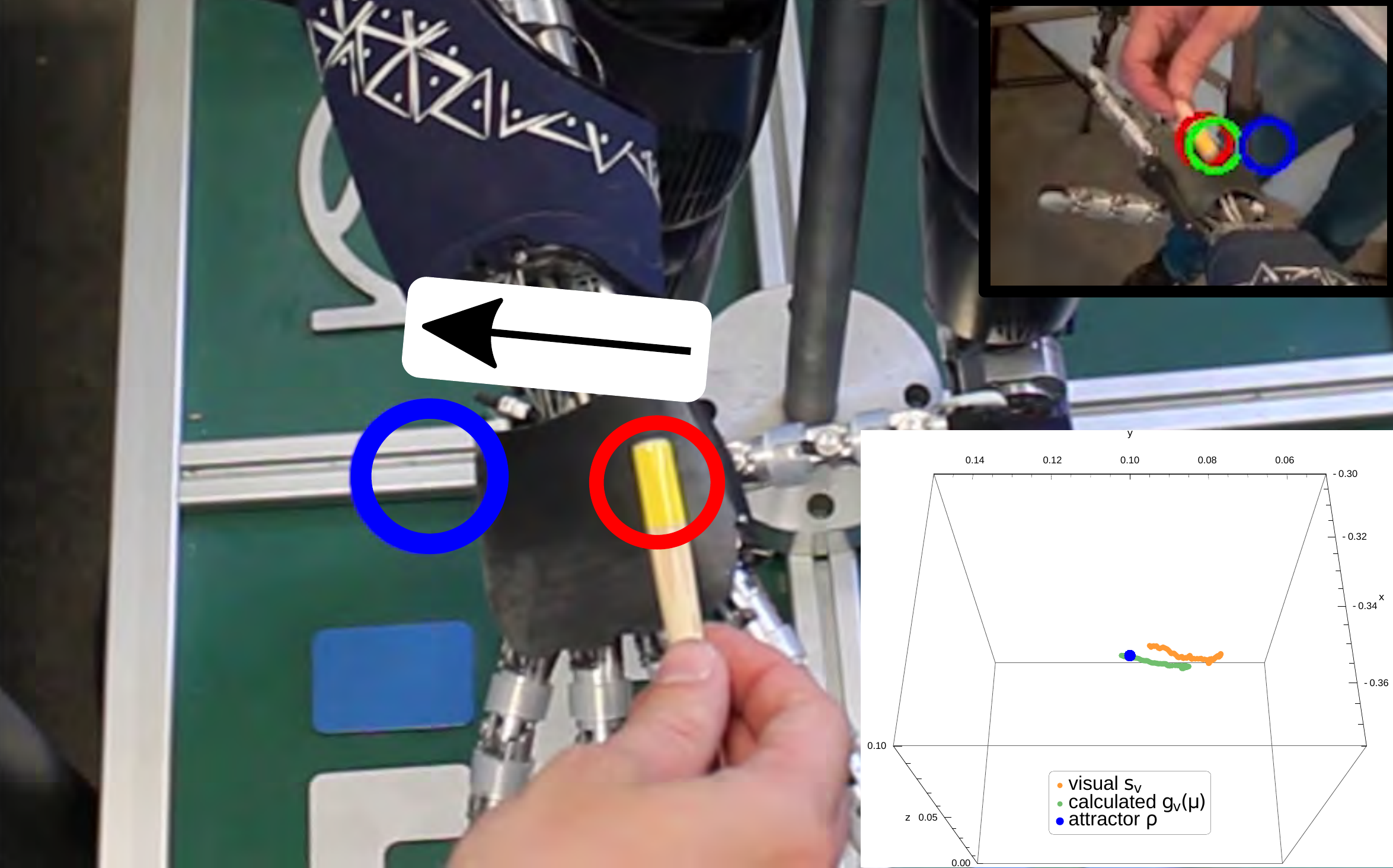}
		\label{results:adapt:right-deviation}
	}	
    \\
	\subfigure[Encoder noise handling path.]{
		\centering
		\includegraphics[width=0.3\textwidth]{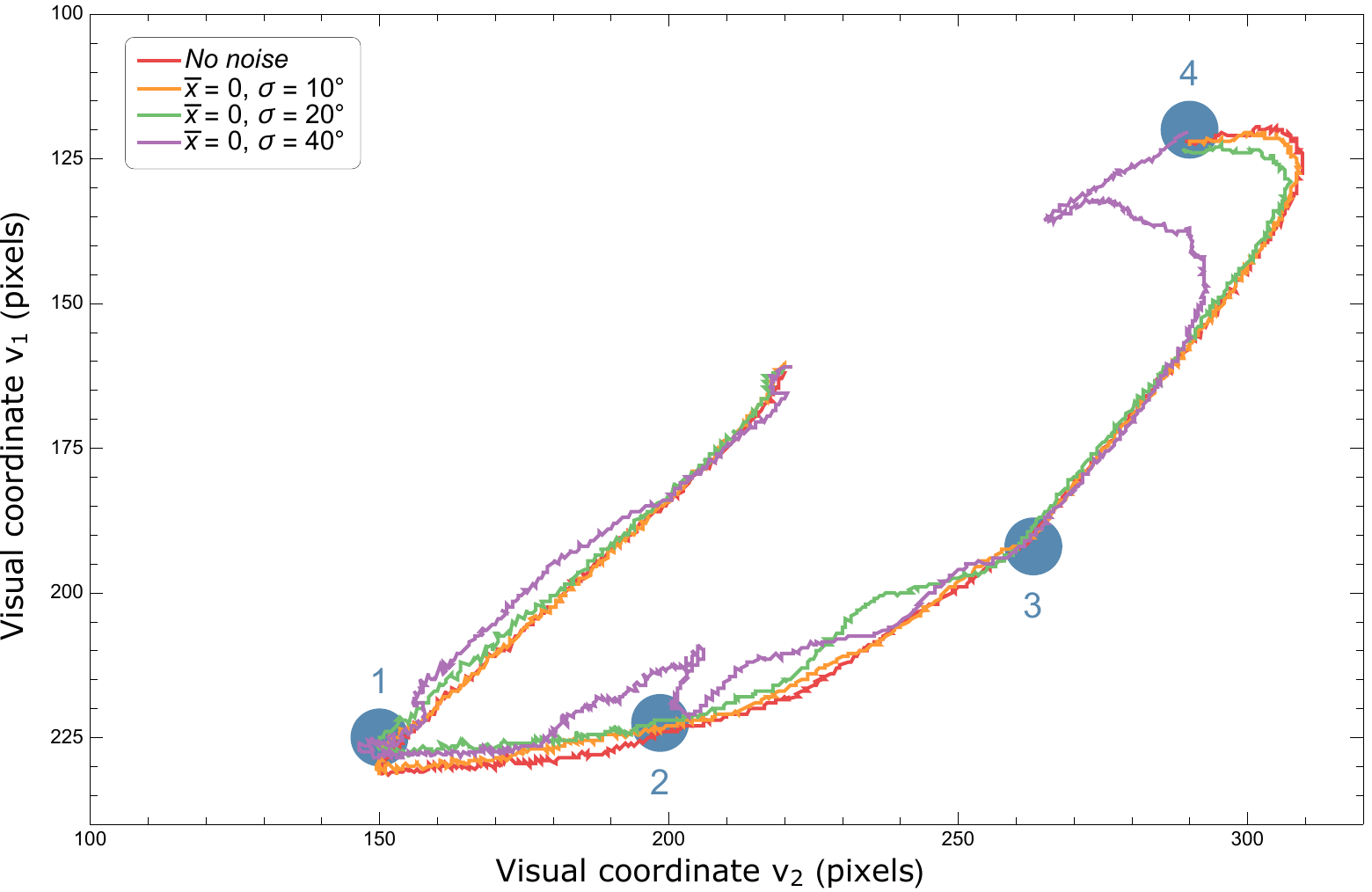}
		\label{results:poc:path-cycle-noise}
	}
	\subfigure[Visual marker located on finger.]{
		\centering
		\includegraphics[width=0.3\textwidth]{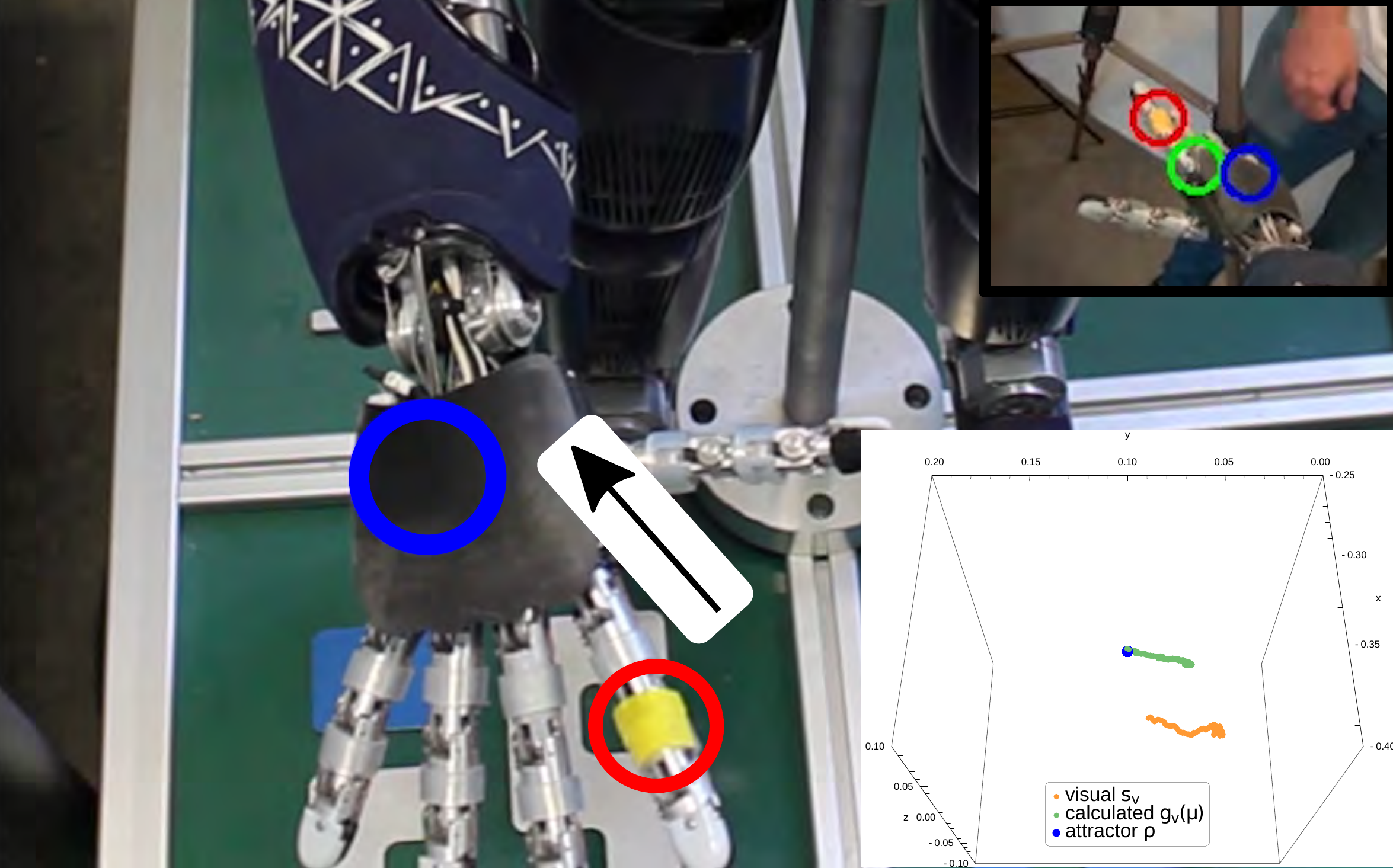}
		\label{results:adapt:finger}
	}	
	\subfigure[Visual marker located on forearm.]{
		\centering
		\includegraphics[width=0.3\textwidth]{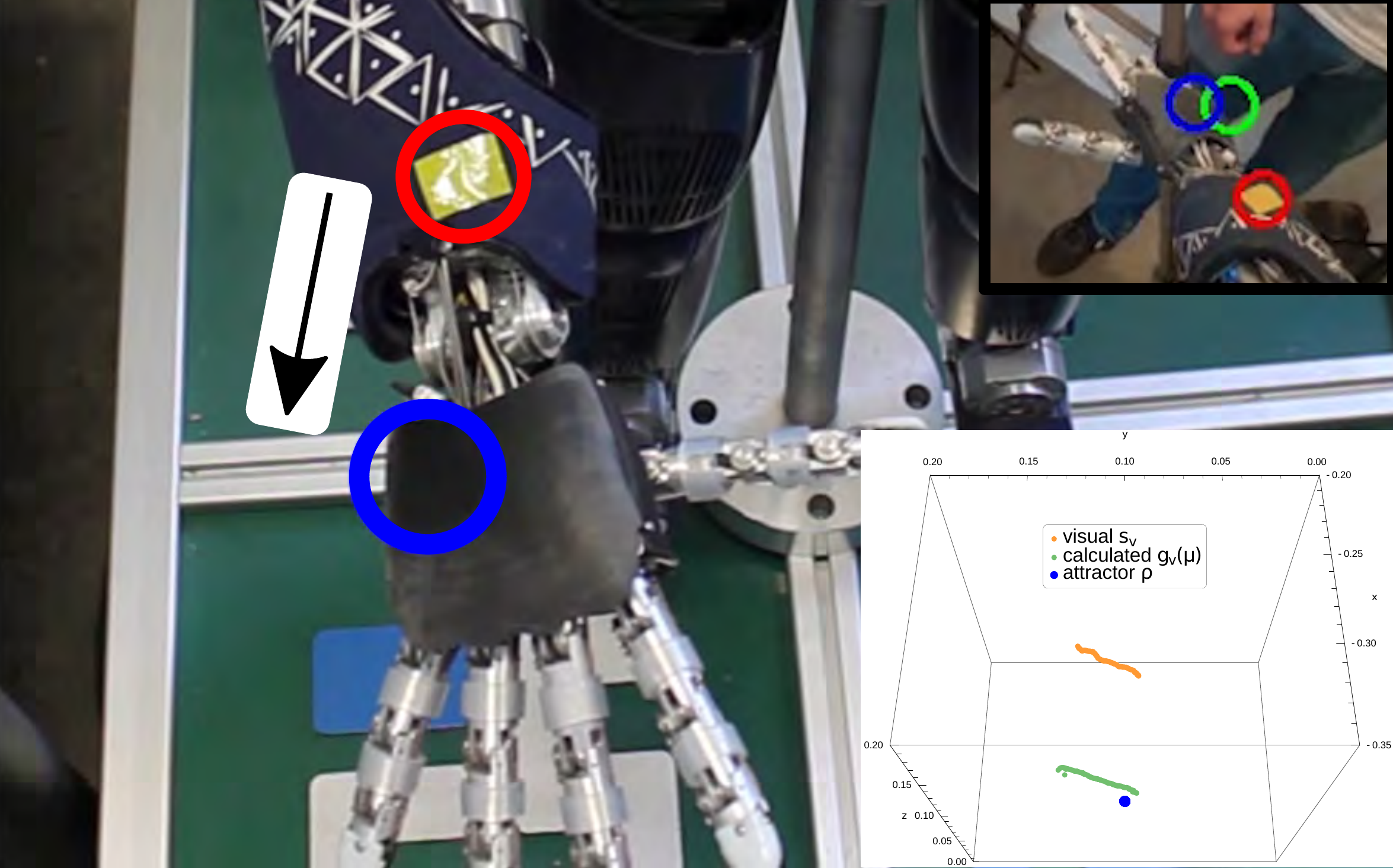}
		\label{results:adapt:forearm}
	}	
	\caption{(Left) End-effector averaged paths in the right arm object reaching for (a) sensory fusion and (d) under noisy encoder sensors. Attractor position is represented by a blue disk. (Right) Results of adaptive behaviour using sensory fusion. If the visual marker position is modified, the algorithm produces a reaching action that tries to approach this new position to the attractor location. On the top right of each image the left eye camera is shown. Attractor position is shown in blue, visual perception in red and calculated position in green. The resulting direction of motion obtained from the algorithm is shown as an arrow.}
	\label{results:pocadapt}
	\vspace{-5mm}
\end{figure*}

To obtain the desired motion, an attractor is defined towards the centre of the left eye image $(c_x, c_y)$. Attractor position $(\rho_5,\rho_6)$ is read from the projected 2D visual input of the left eye and dynamically updates with head motion.

\begin{equation}
\bm{A_e}(\bm{\mu_e},\bm{\rho}) = \rho_4\left(\left[\begin{array}{c} c_x \\ c_y\end{array} \right]  - \left[\begin{array}{c} \rho_5 \\ \rho_6\end{array} \right] \right)
\label{eq:headattractor}
\end{equation}

Internal variable dynamics are then defined in terms of the attractor as: $\bm{f_e}(\bm{\mu_e},\bm{\rho}) = \bm{T_e}(\bm{\mu_e}) \bm{A_e}(\bm{\mu_e},\bm{\rho})$, $\bm{f_e}(\bm{\mu_e},\bm{\rho})\in\rm{I\!R^3}$. With two pixel coordinates and three degrees of freedom, the pseudoinverse of the visual Jacobian matrix is used as the mapping matrix in the visual space: $\bm{T_e}(\bm{\mu_e}) = \bm{J}_{v}^{+}(\bm{\mu_e})$.

\section{Results}
\label{sec:results}
We evaluated the proposed approach for humanoid upper-body reaching in 2D and 3D visual space scenarios. The proposed algorithm, in order to reduce the discrepancy between the current sensory input and the desired state, produces a reaching behaviour with the arm and torso of the robot while the head tracks the object to maintain it in its visual field.


Several experiments were performed: (1) \textit{proof of concept}, right arm reaching towards a series of 2D locations in the visual plane with different levels of noise and sensor sources; (2) \textit{adaptation}, the robot adapts its reaching behaviour during changes on the visual feature location; (3) \textit{comparison}, motion from the active inference algorithm is compared to inverse kinematics; and (4) \textit{dynamics evaluation}, body perception and action variables are analysed during an arm reaching with active head towards a moving object. Video footage of the experiments and some extras can be found in the supplementary video: \url{https://youtu.be/rdbbmwo4TY4}.

The iCub robot is placed in a controlled environment and an easily recognizable object is used as a perceptual attractor to produce the movement of the robot. In 3D space, the values of the causal variables $\rho_1$, $\rho_2$, $\rho_3$ are the coordinates of the 3D position of the object obtained using stereo vision, while in 2D space, the values used for the attractor in \eqref{eq:armattractor} are the object pixel location at the left eye. The value of $\rho_4$ is a weighting factor to adjust the power of the attractor. The right arm end-effector has a visual marker and its 3D position is obtained as the values of $v_1$, $v_2$ and $v_3$, along with its projections in camera space. The relevant parameters of the algorithm are: encoder sensor variance $\Sigma_{s_p}$, visual perception variance $\Sigma_{s_v}$, attractor dynamics variance $\Sigma_{s_\mu}$ and action gains $k_a$. These parameters were tuned empirically with their physical meaning in mind and remained constant during the experiments unless stated otherwise. Their values can be found in the code repository.

\begin{figure*}[t!]
	\centering
	\subfigure[3D position error. Active inference (orange) and inverse kinematics (blue).]{
		\centering
		\includegraphics[width=0.3\textwidth]{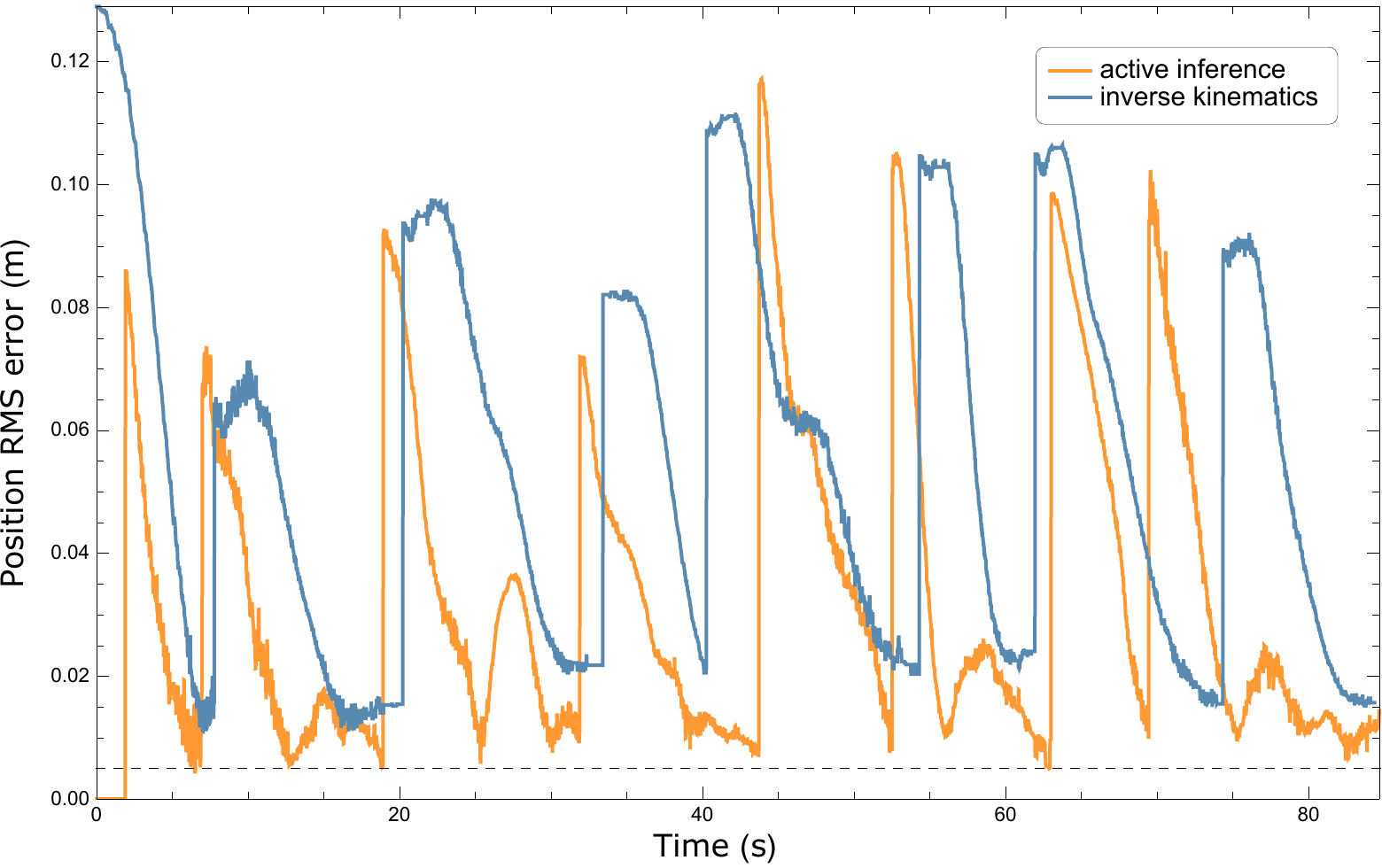}
		\label{results:comp:error-cycle}
	}
	\subfigure[Right arm internal state $\bm{\mu}$ and encoders $\bm{s_p}$ in active inference algorithm.]{
		\centering
		\includegraphics[width=0.3\textwidth]{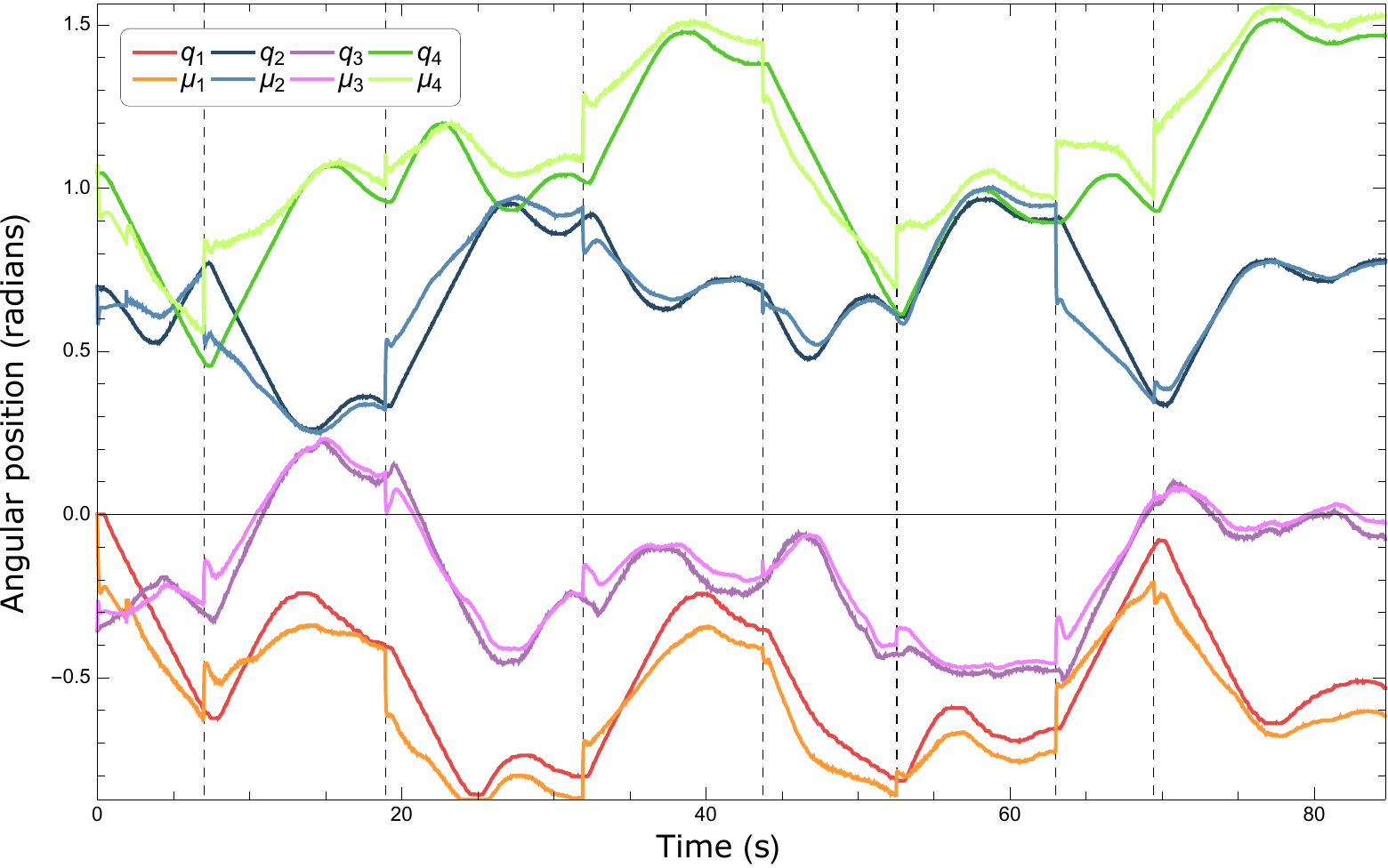}
		\label{results:comp:angle-ai}
	}	
	\subfigure[Right arm encoders $\bm{s_p}$ in inverse kinematics algorithm.]{
		\centering
		\includegraphics[width=0.3\textwidth]{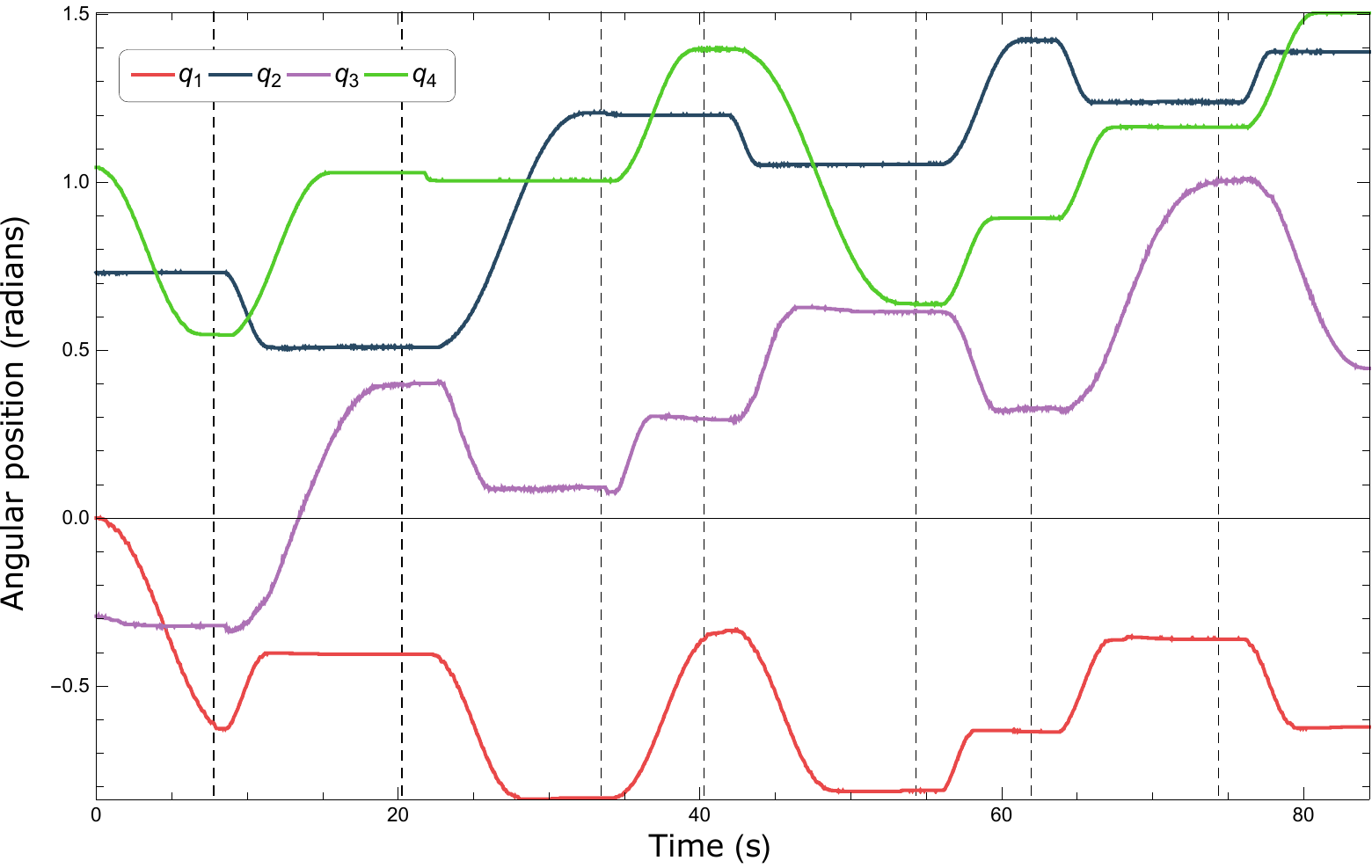}
		\label{results:comp:angle-ik}
	}	
    \\
	\subfigure[Experiment description.]{
		\centering
		\includegraphics[width=0.27\textwidth]{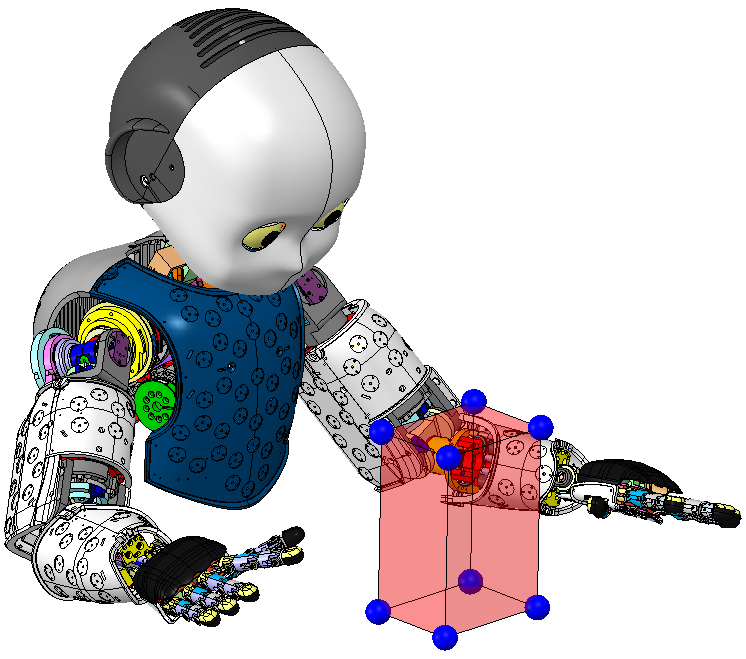}
		\label{results:comp:description}
	}
	\subfigure[3D paths in active inference algorithm.]{
		\centering
		\includegraphics[width=0.27\textwidth]{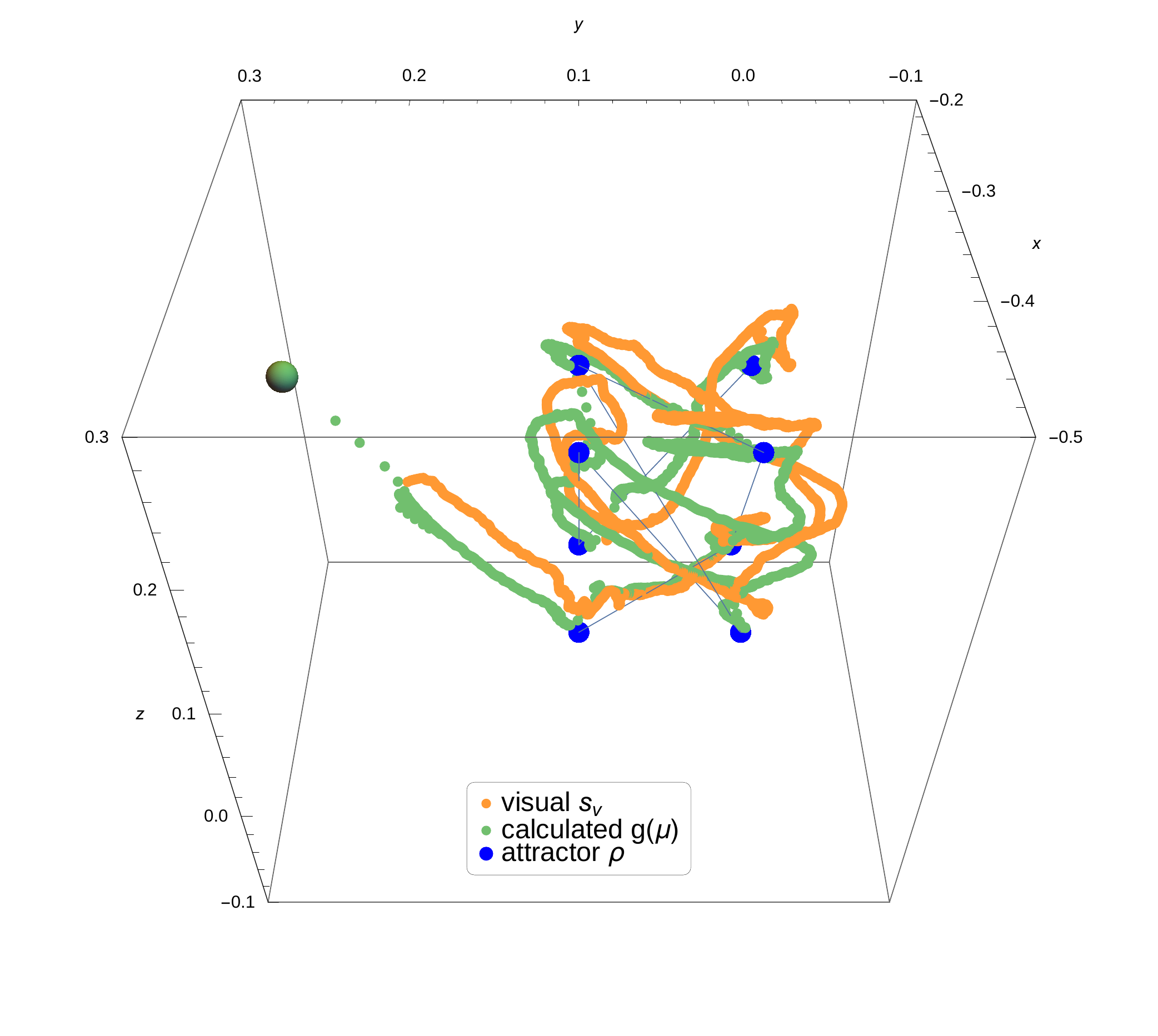}
		\label{results:comp:3d-ai}
	}	
	\subfigure[3D paths in inverse kinematics.]{
		\centering
		\includegraphics[width=0.27\textwidth]{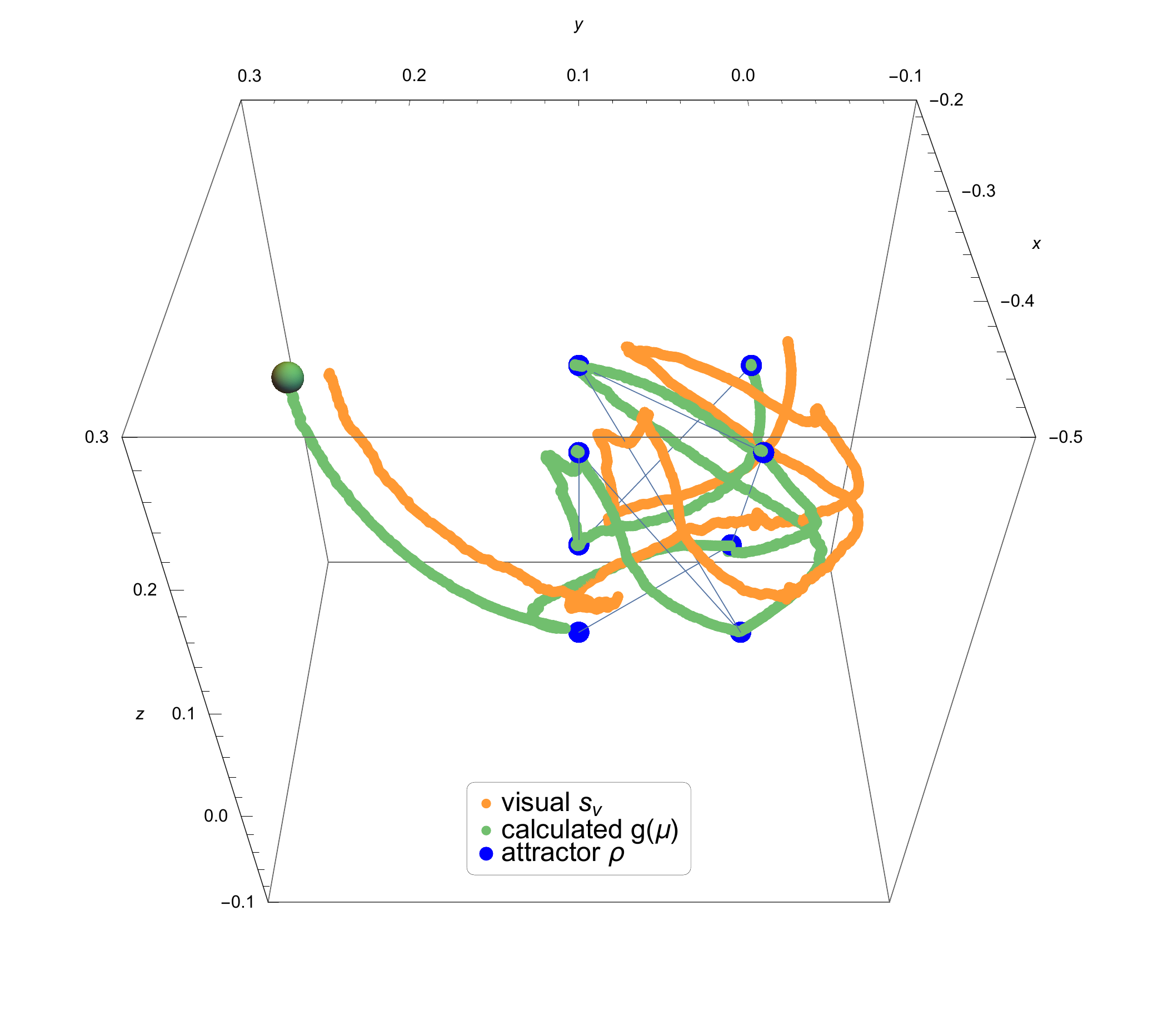}
		\label{results:comp:3d-ik}
	}	
	\caption{Comparison between active inference and inverse kinematics. (Top left) RMS 3D position error between the real visual position and the attractor position. (Top centre and right) Internal state and encoder values of the right arm for each of the algorithms. (Bottom left) Experiment description showing the initial arm position and the rectangular prism. (Bottom centre and right) 3D paths taken for each of the algorithms.}
	\label{results:comparison}
	\vspace{-5mm}
\end{figure*}

\subsection{Sensory fusion and robustness under noisy sensors}
\label{sec:results:poc}

The first experiment was designed to test sensory fusion (Fig. \ref{results:poc:path-cycle}) and the impact of noisy sensors on the algorithm (Fig. \ref{results:poc:path-cycle-noise}). The robot worked in 2D visual space using the left-eye camera (monocular vision) and the torso motion is locked. The robot had to reach 10 times four different static locations in the visual field with the right arm. A location is considered to be reached when the visual position is inside a disk with a radius of five pixels centred at the location position. 


Figure \ref{results:poc:path-cycle} shows the averaged trajectory of the visual marker with different sensor modalities (considering only intrinsic noise from the sensors and model errors): visual, joint angle encoders, and both together. Even though the model has been verified and the camera calibrated, there was a difference between the forward model and the real robot, due to possible deviations in parameters and motor backslash, which implies that the robot had to use visual information to correct its real position. Employing joint angle encoders and vision provided the best behaviour for the reaching of the fixed locations in the visual field, achieving all positions in the shortest fashion. Visual perception reached all the positions but it did not follow the optimum path while using only the encoder values failed to reach all locations.

Secondly, in order to test the robustness against sensory noise, we injected noise in the robot motors encoders $\bm{s_p}$. Four conditions were tested: Gaussian noise with a zero mean and with standard deviation of $\ang{0}$ (control test), $\ang{10}$, $\ang{20}$ and $\ang{40}$. Again, averaged path is shown (Fig. \ref{results:poc:path-cycle-noise}). Trajectories with no noise and $\sigma=\ang{10}$ achieved very similar results, with the first one achieving the objectives slightly faster. Extreme case with $\sigma=\ang{40}$ caused oscillations and erroneous approximation trajectories that produced significant delays in the reaching of the target locations. Intuitively, with low sensor variance $\Sigma_{s_p} < 0.5$ the system was not able to perform the task with high sensor noise.

\subsection{Online adaptive behaviour}
\label{sec:results:adaptation}
We tested the online perception and action loop when there was a change in the visual feature location. The visual feature was in the 3D space, and all degrees of freedom and binocular vision were considered. Figure \ref{results:pocadapt}(b,c,e,f) shows the behaviour of the arm trying to correct the forward model prediction with the altered location of the visual marker. We set the perceptual attractor position to a constant fixed point in 3D space. The end-effector visual marker was removed and replaced with a similar marker located on a tool that was operated by a human. When the marker was moved sideways (Fig. \ref{results:adapt:left-deviation} and \ref{results:adapt:right-deviation}), the algorithm generated an action that moved the end-effector towards the opposite direction to correct the deviation. The internal belief was also updated to find an equilibrium point. Similar behaviour was obtained when the marker was located on one of the fingers (Fig. \ref{results:adapt:finger}) and on the forearm (Fig. \ref{results:adapt:forearm}). In this case, the visual discrepancy with the forward model is much more noticeable and the algorithm is not able of fully reaching the visual marker position to the attractor, but the same opposite action behaviour as in the previous case is obtained.

This experiment shows that in the case of visual sensorimotor conflict a perception body alteration and action to reduce the uncertainty is generated. In essence, this is a simplified body illusion that only uses one visual feature. Thus, this is extremely relevant as the model of the robot is predicting drifts on the hand localization \cite{hinz2018drifting} but also actions.

\begin{figure*}[t!]
	\centering
	\subfigure[Right arm internal states $\bm{\mu}$ and encoders $\bm{s_p}$.]{
		\centering
		\includegraphics[width=0.3\textwidth, height=95px]{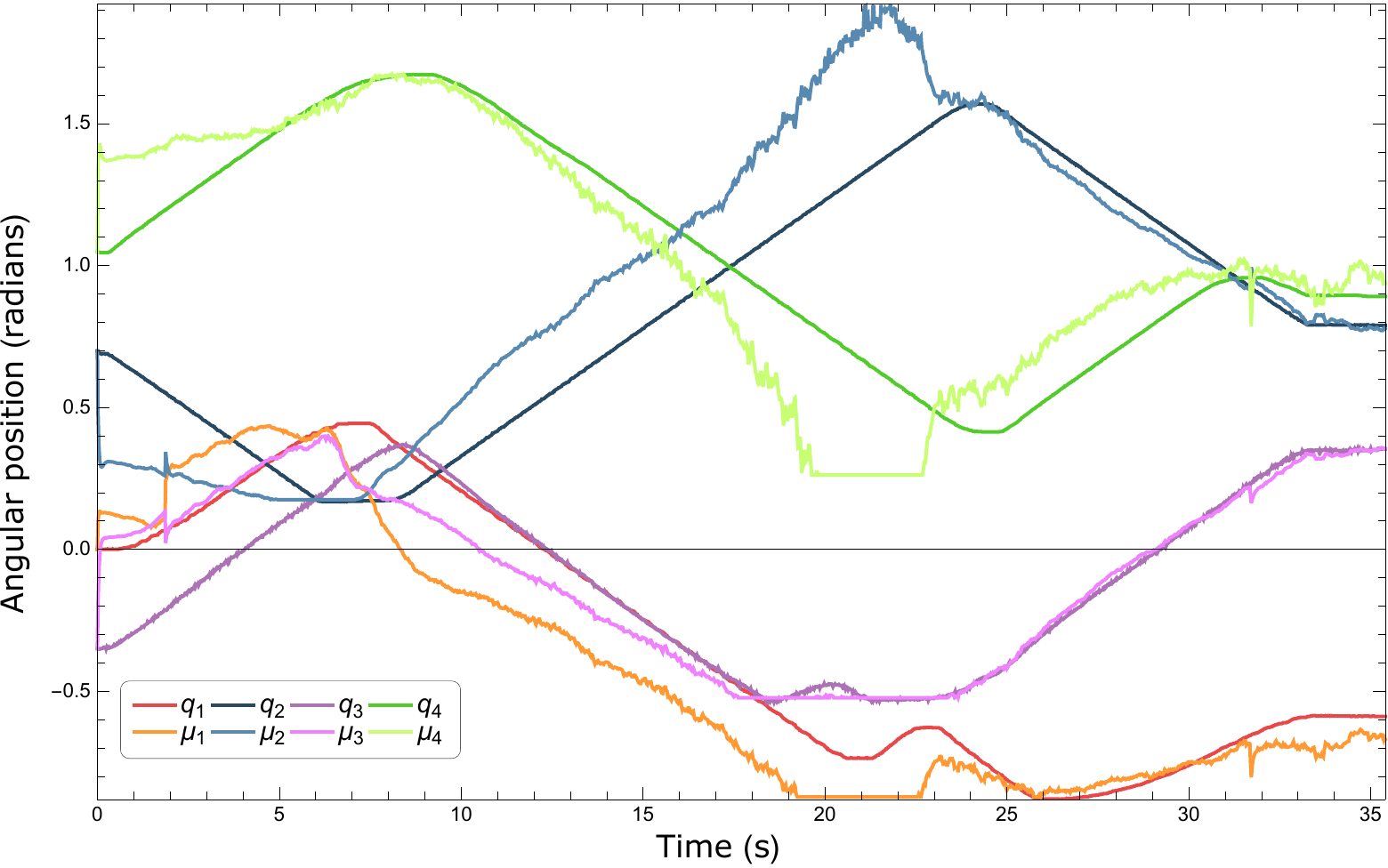}
		\label{results:dynamics:encoders}
	}
	\subfigure[Head internal states $\bm{\mu_e}$ and encoders $\bm{s_e}$.]{
		\centering
		\includegraphics[width=0.3\textwidth, height=95px]{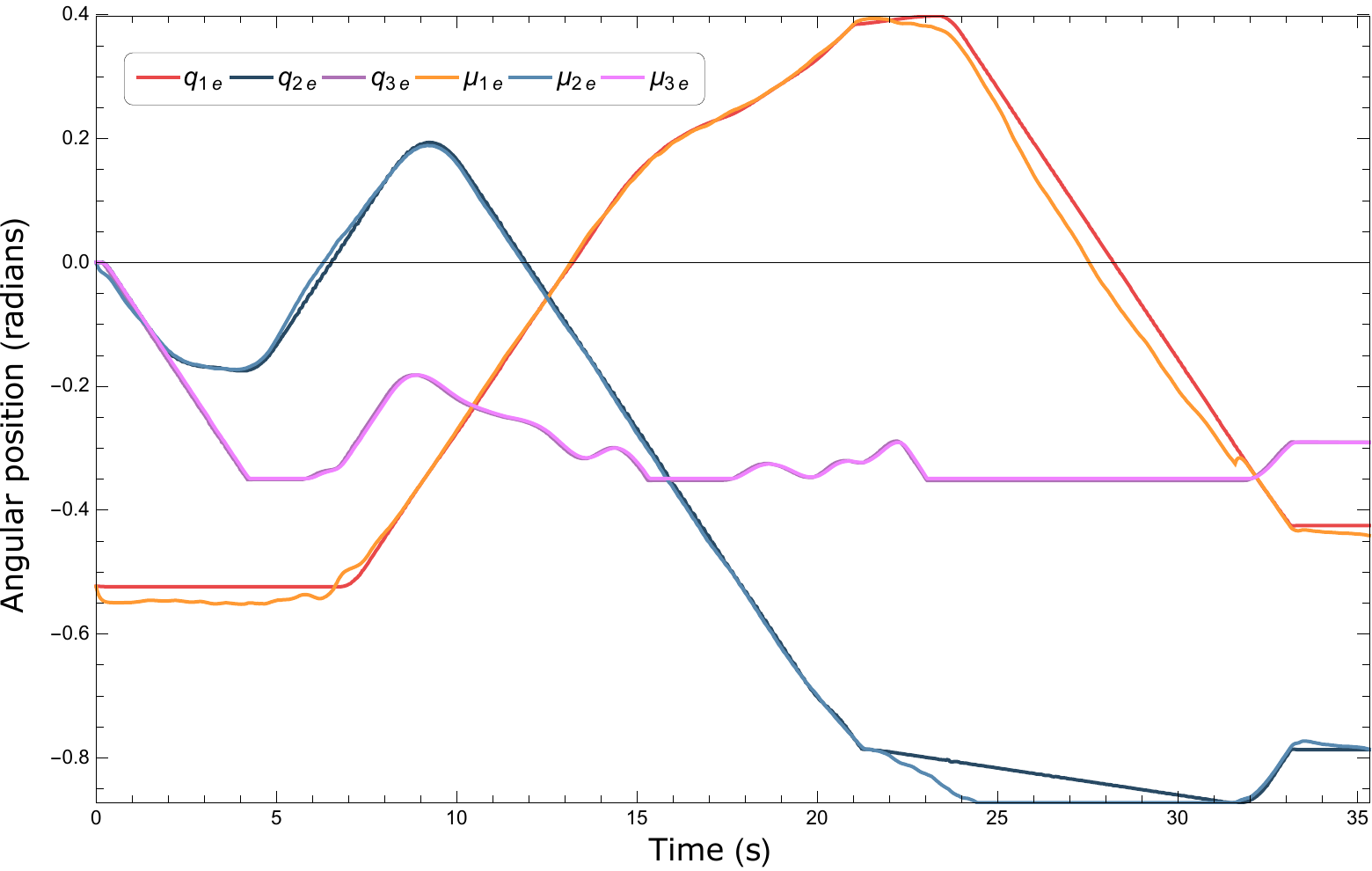}
		\label{results:dynamics:encoders:head}
	}	
	\subfigure[3D paths of $\bm{s_v}$, $\bm{g_v(\mu)}$ and $\bm{\rho}$.]{
		\centering
		\includegraphics[width=0.3\textwidth, height=95px]{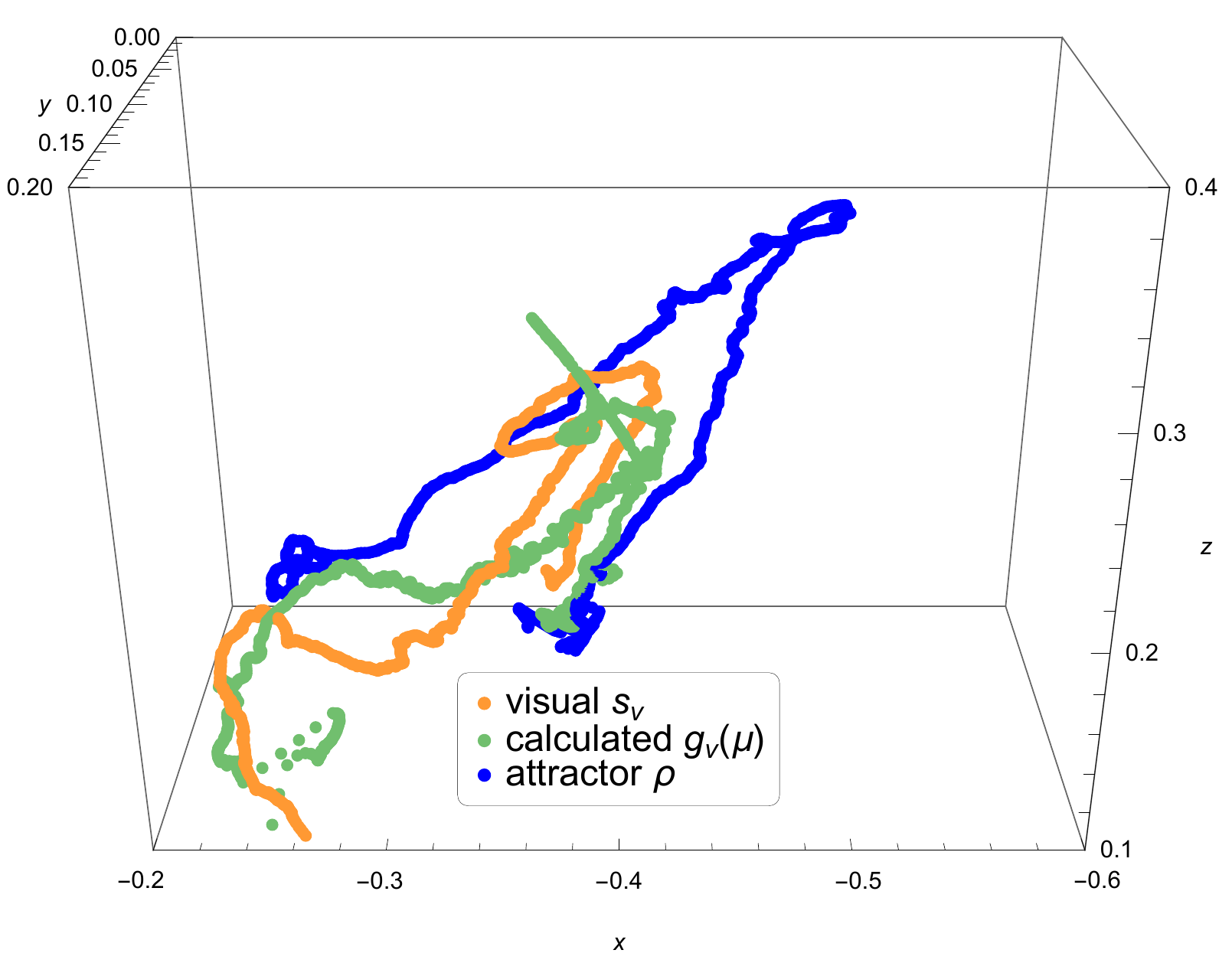}
		\label{results:dynamics:3Dpath}
	}	
    \\
	\subfigure[Right arm joint error $(\bm{s_p} - \bm{\mu})$.]{
		\centering
		\includegraphics[width=0.3\textwidth, height=95px]{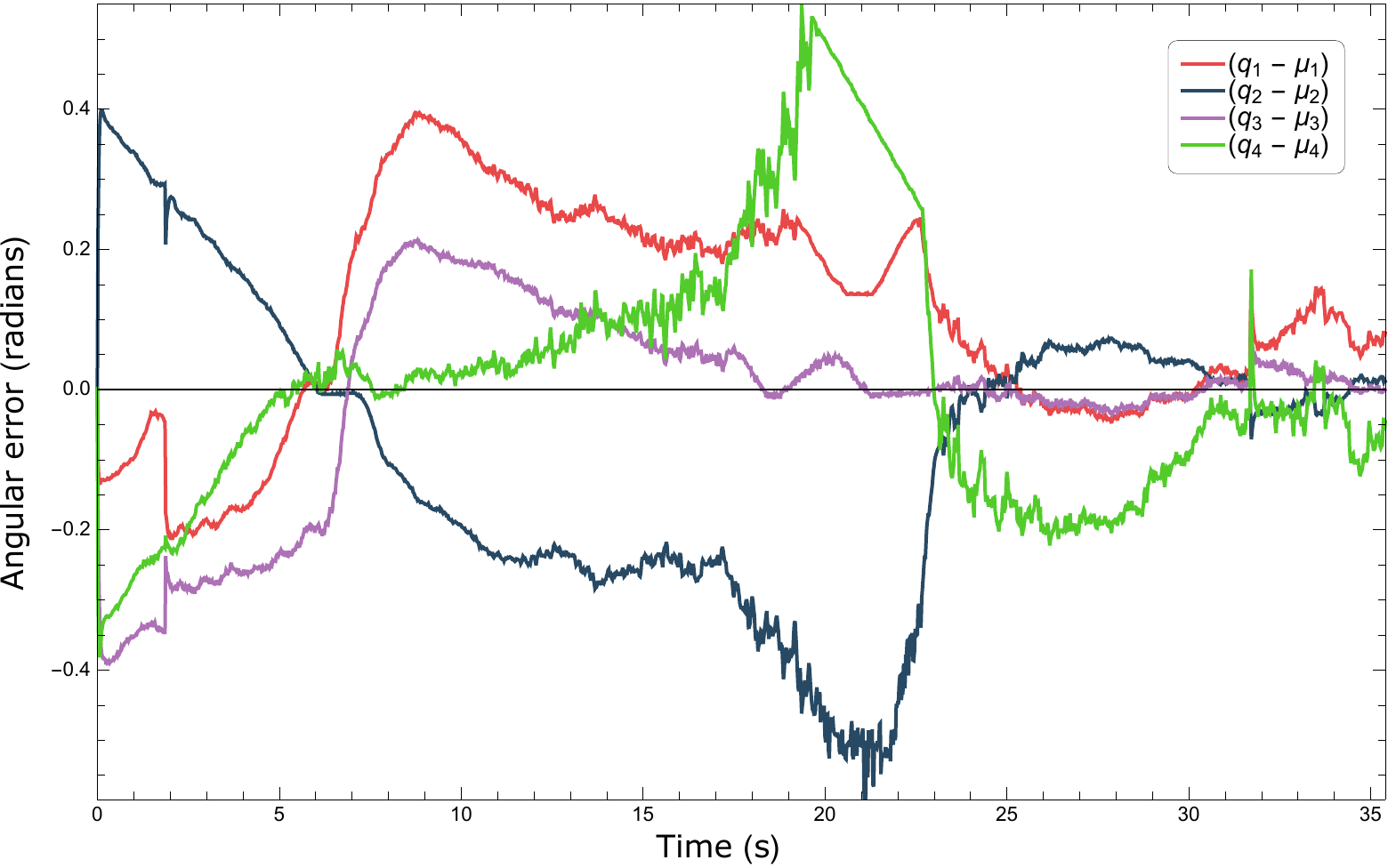}
		\label{results:dynamics:error-rightarm}
	}
	\subfigure[Head joint error $(\bm{s_p} - \bm{\mu})$.]{
		\centering
		\includegraphics[width=0.3\textwidth, height=95px]{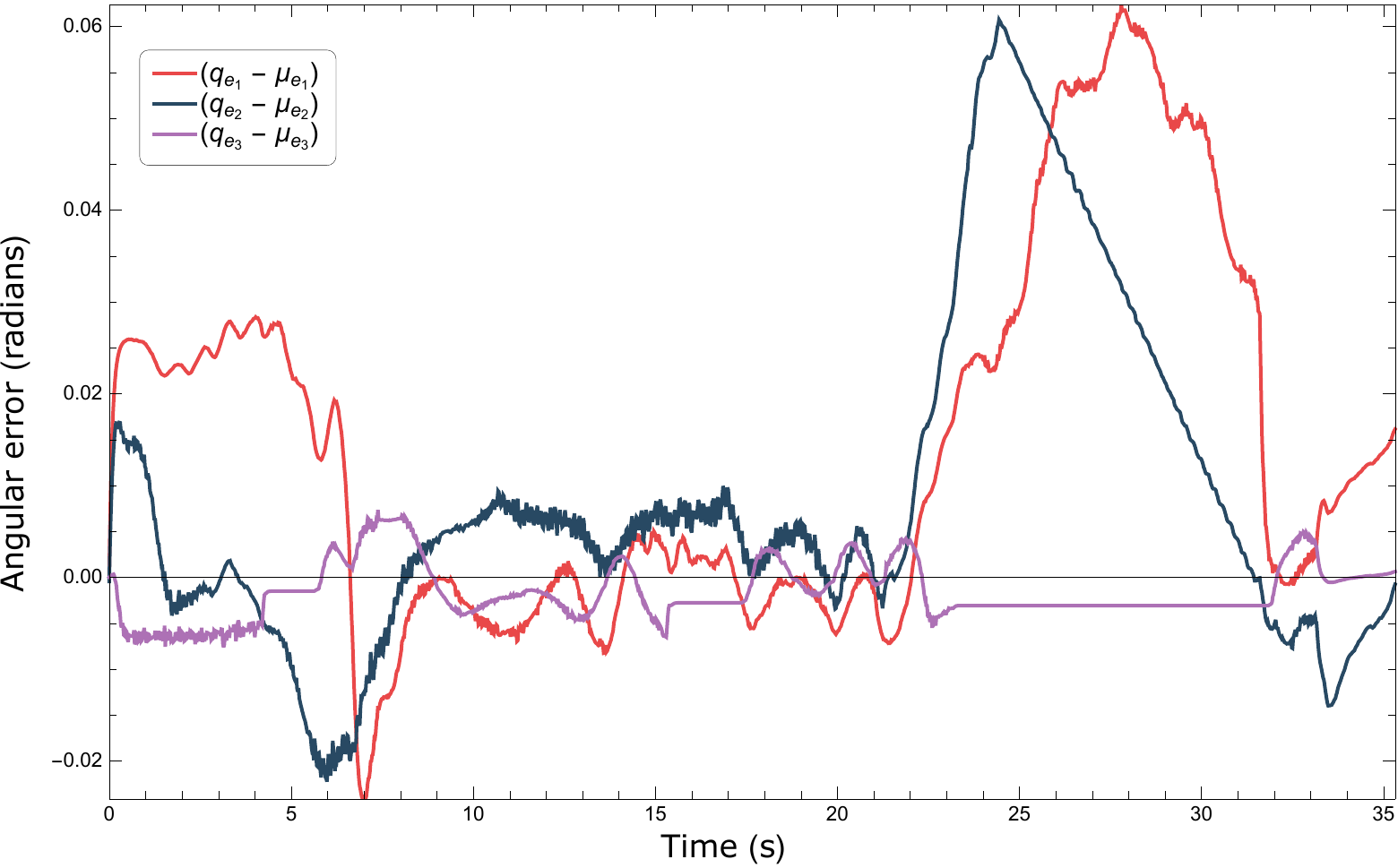}
		\label{results:dynamics:error-head}
	}	
	\subfigure[3D end-effector error $(\bm{s_v} - \bm{g_v}(\bm{\mu}))$.]{
		\centering
		\includegraphics[width=0.3\textwidth, height=95px]{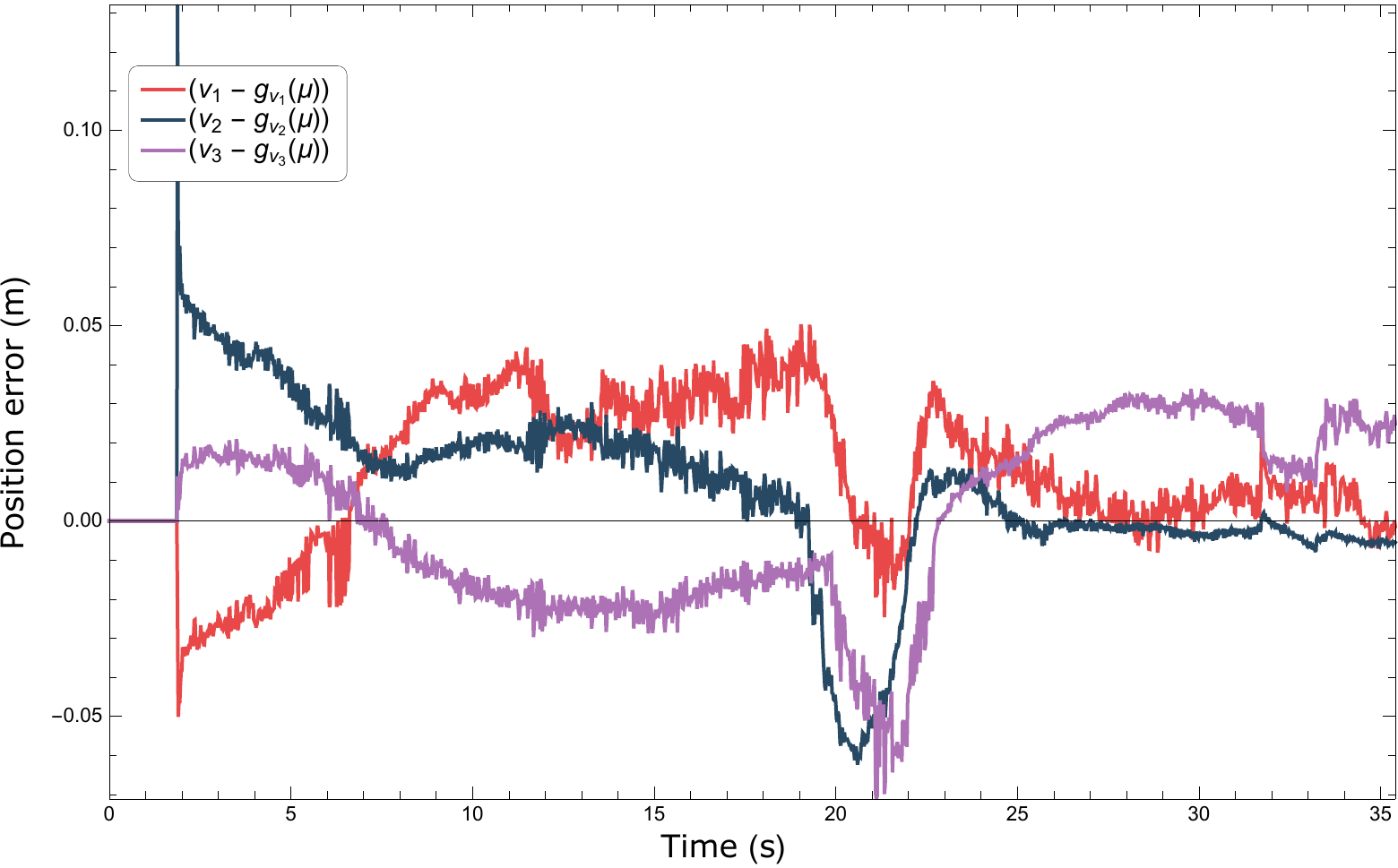}
		\label{results:dynamics:error-3D}
	}

	\caption{System dynamics for reaching a moving object in 3D. Internal states are driven towards the perceptual attractor position. The difference between the value of internal states and encoders (\ref{results:dynamics:error-rightarm}) and between 3D calculated position and visual location of end-effector (\ref{results:dynamics:error-3D}) drive the actions. 
	}
	\label{results:dynamics}
	\vspace{-5mm}
\end{figure*}

\subsection{Comparison of active inference and inverse kinematics}
\label{sec:results:comparison}

We compared our algorithm with the inverse kinematics approach in a reaching task using the right arm, the torso and the head. This comparison was performed by defining a set of 3D waypoints matching the vertices of a rectangular prism that the robot must reach, as described in Fig. \ref{results:comp:description}.

The inverse kinematics module from iCub\footnote{\textit{iKin} uses the nonlinear optimization package \textit{Ipopt} to obtain the joint angle position to arrive at a certain 3D point in space taking a certain angle configuration as the starting point.} was used sequentially to solve each of the eight vertices reaching problems. The robot joints were position-controlled to the desired state and joint speed was adjusted in order for the motion to have the same time duration as the active inference test. Head angles were positioned in a way such that the perceptual attractor was always inside the visual field. Our approach included in addition to the arm and the torso, also active head object tracking. In this case, the head started looking front and the arm was outside of the visual field. 



Each location was considered to be reached once the calculated position of the visual marker was inside a sphere of $r = 0.5\,cm$ centred at the attractor location. Setting the 3D position reconstructed from the visual input as the ground truth for the location of the end-effector, the performance was assessed using the root mean square (RMS) error between this location and the target position of each vertex. Figure \ref{results:comp:error-cycle} shows the RMS error between the visual perception position $\bm{s_v}$ and the attractor position $\bm{\rho}$. The dashed line marks an error of $0.5\,cm$. Because the active inference algorithm implements online sensory fusion (forward model and stereo vision) the performance of the reaching task was improved over the inverse kinematics approach. The minimum error obtained for the inverse kinematics algorithm was $1.1\,cm$, while the minimum error for the active inference algorithm was $0.42\,cm$. The mean error for the reaching of all 3D points in the inverse kinematics was $\overline{e}_{IK} = 1.69\,cm$, while the active inference algorithm obtained $\overline{e}_{AI} = 0.67\,cm$.

Figure \ref{results:comp:angle-ai} shows the internal state belief $\bm{\mu}$ and the encoder values $\bm{q}$ of the active inference algorithm, while Fig. \ref{results:comp:angle-ik} shows the encoder values $\bm{q}$ for the inverse kinematics algorithm. The attractor position change is represented using a dashed line. In the active inference algorithm, each time the attractor position changes the believed state $\bm{\mu}$ gets pushed towards the new position and there is a relatively big discrepancy between $\bm{\mu}$ and $\bm{q}$ that gets reduced as the goal is reached. 


Figures \ref{results:comp:3d-ai} and \ref{results:comp:3d-ik} show the paths taken by the active inference and the inverse kinematics algorithm. In both algorithms, the robot reached all targets, but there was a clear difference in the discrepancy between the calculated and the visual perception position. Inverse kinematics paths show the error between the model and the physical robot (distance between the orange and the green line of Fig. \ref{results:comp:3d-ik}). Our approach reduced this distance when approaching the target location in a closed-loop manner (Fig. \ref{results:comp:3d-ai}), while inferring its real body configuration and using head object tracking.

\subsection{System dynamics when reaching a moving object}
\label{sec:results:dynamics3D}

We evaluated the algorithm in a reaching task of a moving object manually operated by a human. The experiment started with the object near the robot and it was progressively moved away and in an ascendant vertical direction, to finally be handed to the robot in an intermediate and lower location. 

\begin{figure}[hbtp!]
	\centering
	\subfigure{
		\includegraphics[width=0.45\columnwidth, height=90px]{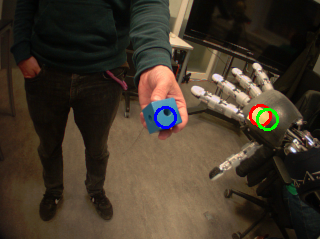}
		\label{results:dynamics2D:vision}
	}	
	\subfigure{
		\includegraphics[width=0.48\columnwidth, height=90px]{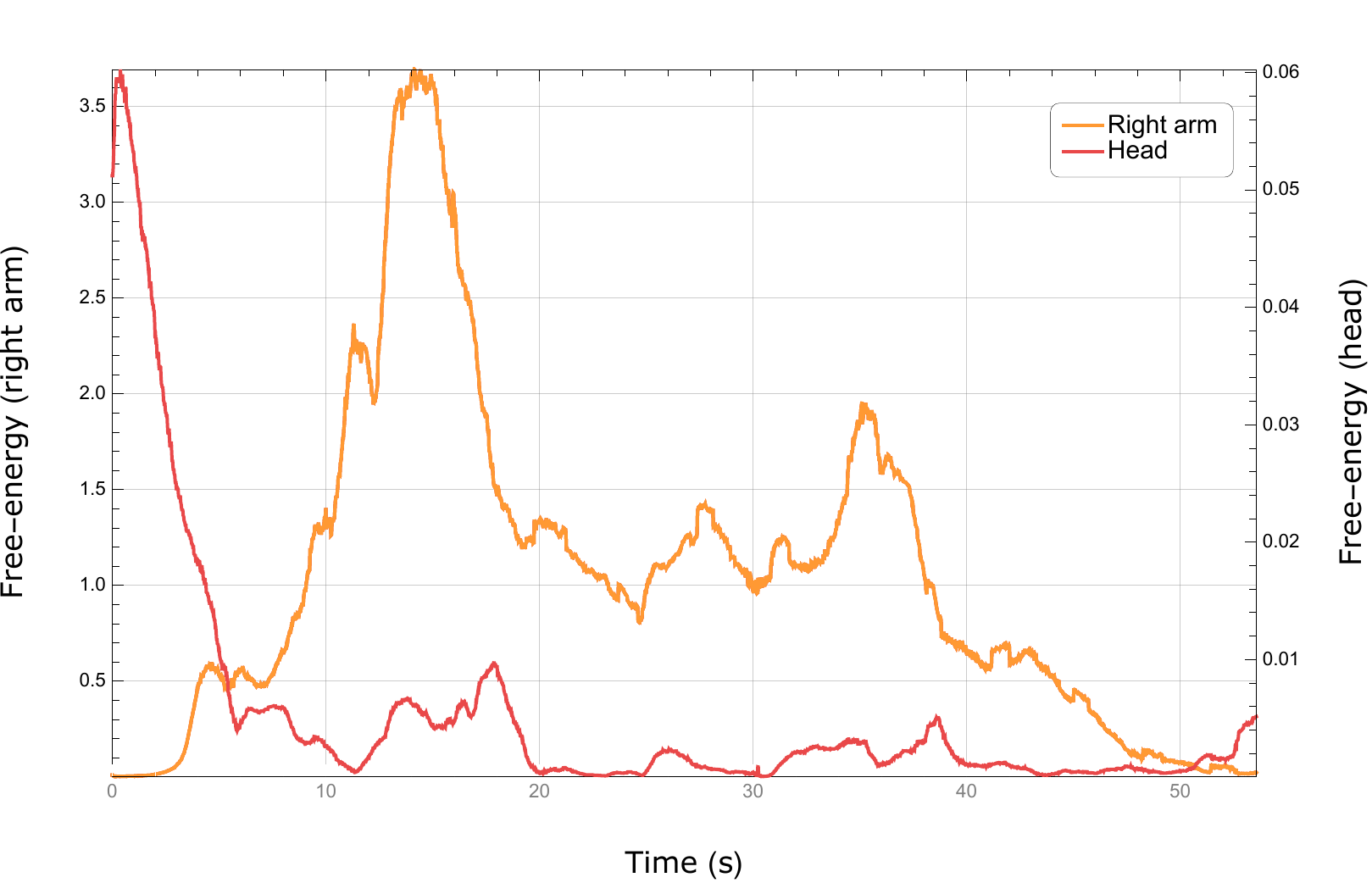}
		\label{results:dynamics2D:fe}
	}
	\caption{Reaching a moving object in 2D. (Left) Robot vision: visual feature (red), predicted end-effector location (green), object (blue). (Right) Free energy optimization for the arm (orange) and the head (red).}
	\label{results:dynamics2D}
\end{figure}

This experiment was performed both in 2D and 3D visual space. In 2D space, the setup used monocular vision from the left eye and the torso was blocked. Our model was tested for both arms and using active head object tracking. In 3D space, we enabled the torso motion but restricted the model for the right arm and the head. Figure \ref{results:dynamics2D} shows the vision of the robot and the free energy values for the head and the arm during the reaching experiment in 2D space. $F$ is minimised modifying the arm perception and generating actions towards the object. 


The resulting variable dynamics of the 3D space experiment are shown in Fig. \ref{results:dynamics}. The initial position of the right arm lied outside of the visual plane (Fig. \ref{results:dynamics:error-3D} showing zero error). When there was no visual input, the free energy optimization algorithm only relied on joint measurements and the forward model to produce the reaching motion until the right hand appears in the visual plane. Figures \ref{results:dynamics:encoders} and \ref{results:dynamics:encoders:head} show both the encoders measurements $\bm{q}$ and the estimated joint angle $\bm{\mu}$ of the arm and head. Figure \ref{results:dynamics:3Dpath} shows how calculated $\bm{g_v(\mu)}$ and real $\bm{s_v}$ 3D positions of the right arm end-effector follow the perceptual attractor $\bm{\rho}$. Although the cameras were previously calibrated, noise was present in the stereo reconstruction of visual perception. This noise was more relevant in the depth estimation (first term in Fig. \ref{results:dynamics:error-3D}). Stop action for this experiment was produced by the sense of touch. Contact in the hand pressure sensors triggers the grasping motion.


\section{Active inference in real artificial agents}
\label{sec:discussion}

Being able to develop robots with biologically plausible body perception and action models allow us to validate current brain-inspired computational models \cite{lanillos2017enactive,hoffmannrobots}. However, there is a big gap between the theoretical mathematical construct and physical reality. This work reduces this gap and shows the plausibility of implementing the free energy principle \cite{friston2010unified} on artificial agents for real-world applications, joining together robotics and computational neuroscience. The reproduced robot behaviour, during the system dynamics study (Sec. \ref{sec:results:dynamics3D}), resembled a one-year-old infant. Just by prediction error minimization, in this case, through surprise minimization, the robot was able to perform human-like movements in a reaching and object tracking task. The model and the experiments can be reproduced and replicated by downloading the open-source code 
\url{tobereleased}.

\subsection{Relevant model predictions for human body perception}

Interestingly, our model predicts the appearance of involuntary actions in a sensorimotor conflict situation, such as the rubber-hand illusion. In the presence of sensory errors (arm location expectation mismatch), we observed movements towards the visual feature of the arm. In previous work, we showed that the proprioceptive drift (mislocalization of the hand) using a similar free energy optimization algorithm could be modelled \cite{hinz2018drifting}. By adding the action term of active inference, forces towards minimising the prediction error should appear, i.e., in the same direction of the drift. This behaviour could be in charge of the online adaptation of movements in the presence of disturbances or not trained conditions, bypassing the voluntary movement pathway.

\subsection{Technical challenges for a full-fledged body model}
Closed-loop adaptive perception and action is an important advantage. The model is assumed to be an approximation of the reality and the minimization of the variational free energy tackles both sensory noise and model bias. In fact, the equilibrium point where the algorithm converges reduces the differences between the model and the real observations. However, it is prone to local minima. As there are no task hierarchies, dual arm reaching produced redundant motions in both arms. A hierarchical version of the proposed approach, as proposed in \cite{tani2003learning}, could circumvent this limitation by generating attractors for specialised behaviours.

The engineering analysis needed on iCub specific parameters and actuation was the biggest limitation for generalization. Sensor variances, action gains and velocity limits were experimentally tuned. In theory, variances and gains could be also optimised within the free energy framework \cite{friston2010action}. Besides, we introduced how time discretization in a velocity control scheme can be used to simplify the action computation. In force controlled non-linear systems the mapping between the sensory consequence and the force applied must be learnt \cite{lanillos2018active} or explicitly computed through the forward dynamics \cite{pio2016active}, considerably increasing the complexity of the optimization framework.

The proposed algorithm is scalable to integrate features from different modalities (i.e., adding a new term to the summation) and handles sensory fusion in both perception and action. Furthermore, the needed generative functions can be easily computed from the forward kinematics or be even replaced by learned ones, as shown in \cite{lanillos2018adaptive}. However, more complex sensory input features should be incorporated to be able to compare with other bio-inspired approaches such as deep reinforcement learning.





\section{Conclusions}
\label{sec:conclusions}
This work presented a neuroscience-inspired closed-loop algorithm for body perception and action working in a humanoid robot. Our \textit{active inference} model is the first one validated on a real robot for upper-body reaching and active head object tracking. Robots perceiving and interacting under the active inference approach is an important step for evaluating the plausibility of the model in biological systems as we systematically observe the behaviours and connect them to the mechanisms behind, and even make predictions. As a proof of concept, we showed its robustness to sensory noise and discrepancies between the robot model and the real body, its adaptive characteristic to visual online changes and its capacity to embed sensory fusion. Body perception was approached as a hidden Markov model and the unobserved body configuration/state was continuously approximated with visual and proprioceptive inputs by minimising the free energy bound. The action was modelled within perception as a reflex to reduce the prediction error. We tested the algorithm on the iCub robot in 2D and 3D using binocular vision. Results showed that our approach was capable of combining different independent sources of sensory information without an increase of computational complexity, displaying online adaptation capabilities and performing more accurate reaching behaviours than inverse kinematics





\bibliographystyle{IEEEtran}
\bibliography{ninaRHI,pl,selfception,icub}

\end{document}